\documentclass[10pt,twocolumn,letterpaper]{article}

\usepackage{cvpr}              % To produce the CAMERA-READY version
\usepackage{graphicx}
\usepackage{amsmath}
\usepackage{amssymb}
\usepackage{booktabs}

\usepackage[pagebackref,breaklinks,colorlinks]{hyperref}

\usepackage[capitalize]{cleveref}
\crefname{section}{Sec.}{Secs.}
\Crefname{section}{Section}{Sections}
\Crefname{table}{Table}{Tables}
\crefname{table}{Tab.}{Tabs.}

\usepackage[labelsep=period]{caption}
\usepackage{xcolor}
\usepackage{enumitem}

\renewcommand{\paragraph}[1]{\vspace{0.2em}\noindent \textbf{#1 \hspace{0.2em}}}

\definecolor{MyDarkRed}{rgb}{0.66, 0.16, 0.16}
\definecolor{MyDarkBlue}{rgb}{0.16, 0.16, 0.66}
\usepackage{lipsum} % Just for generating dummy text, can be removed

\usepackage[marginal]{footmisc}

\usepackage{url}
\usepackage{colortbl}
\usepackage[accsupp]{axessibility}

\begin{document}

%\title{Geometric Contrastive Masking Modeling for 3D Scene Understanding}
% \title{MM-3DScene: Customizing Masked Modeling for 3D Scene Understanding}
\title{
MM-3DScene:
3D Scene Understanding by Customizing Masked Modeling with Informative-Preserved Reconstruction and Self-Distilled Consistency
% MM-3DScene: Informative Areas Preserved Masked Modeling with 
%  Consistency Distillation for 3D Scene Understanding 
%   \mt{MM-3DScene: Customizing Masked Modeling for 3D Scene Understanding by Preserving Informative Areas and Self-Distilling Consistency}
%     \mt{or MM-3DScene: Customizing Masked Modeling for 3D Scene Understanding}
}

% \title{Exploring Masked Modeling and Unifying it with Contrastive Learning for 3D Scene Understanding}
\author{
Mingye Xu\textsuperscript{\rm 1,3,4}\footnotemark[1],
Mutian Xu\textsuperscript{2,5}\footnotemark[1],
Tong He\textsuperscript{4},
Wanli Ouyang\textsuperscript{4},
Yali Wang\textsuperscript{1,4}\footnotemark[2],
Xiaoguang Han\textsuperscript{2,5},
Yu Qiao\textsuperscript{1,4}\footnotemark[2] 
\vspace{5pt}\\
	\normalsize{\textsuperscript{\rm 1} The Guangdong Provincial Key Laboratory of Computer Vision and Virtual Reality Technology,} \\
 \normalsize{Shenzhen Institute of Advanced Technology, Chinese Academy of Sciences, Shenzhen 518055, China}\\ %If you have multiple authors and multiple affiliations
 	\normalsize{\textsuperscript{\rm 2} SSE, CUHKSZ} \qquad \normalsize{\textsuperscript{\rm 3}University of Chinese Academy of Sciences}\\
	 %If you have multiple authors and multiple affiliations
	\normalsize{\textsuperscript{\rm 4}Shanghai Artificial Intelligence Laboratory} \qquad \normalsize{\textsuperscript{\rm 5} FNii, CUHKSZ}\vspace{5pt}\\
 \small{\href{https://mingyexu.github.io/mm3dscene/}{mingyexu.github.io/mm3dscene}}
}

% \and

% \maketitle

\twocolumn[{
    \renewcommand\twocolumn[1][]{#1}
    \maketitle
    \begin{center}
        \centering
        \captionsetup{type=figure}
        \includegraphics[width=1\textwidth]{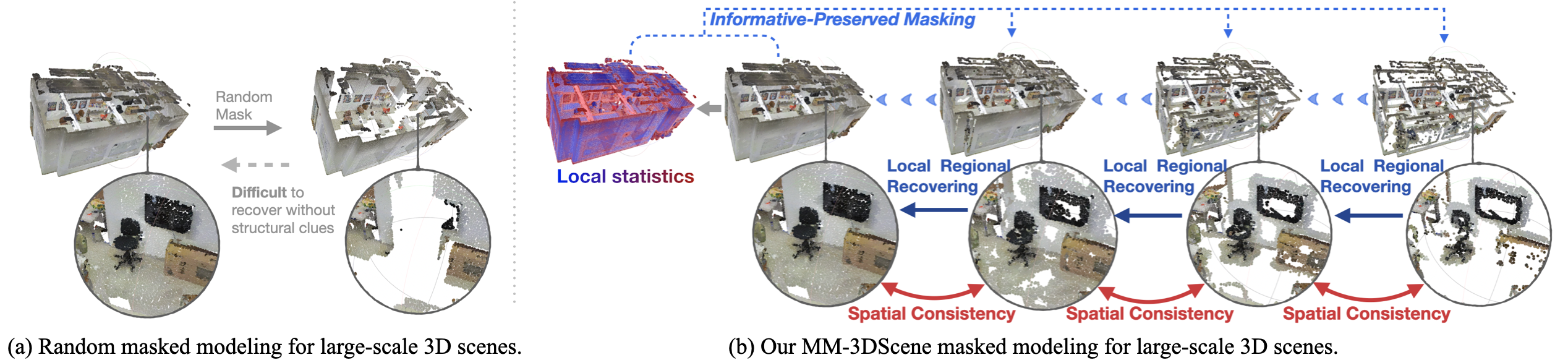}
        \captionof{figure}{
        \textbf{\textit{How to apply masked modeling for large-scale 3D scenes?}} 
        (a) Conventional random masked modeling on 3D scenes may cause a high risk of uncertainty.In this figure, a \textbf{\textit{chair}} and a \textbf{\textit{TV}} are totally masked, which are extremely difficult to be recovered without any context guidance. \textbf{(b)} Our \textbf{MM-3DScene} exploits local statistics to discover and \textit{preserve} representative structured points, effectively simplifying the pretext task.
        At each learning step, our method focuses on restoring \textit{regional} geometry, and enjoys less ambiguity. Moreover, since unmasked areas are underexplored during reconstruction, the model is encouraged to maintain the intrinsic \textit{spatial consistency} on unmasked points between different masking ratios, which requires the consistent understanding of unmasked areas.
        } \label{fig:intro}
    \end{center}
}]

\footnotetext[1]{Equal contribution.}
\footnotetext[2]{Corresponding authors.}

% \begin{figure*}[!h] \centering
%     \includegraphics[width=0.9\textwidth]{imgs/fig_intro_1.png}
%     \caption{Teaser image. Usually, we will have a teaser image on the first page. It can be the key idea of the proposed method or some eye-catching results.} \label{fig:information_density}
% \end{figure*}

%%%%%%%%% ABSTRACT
\begin{abstract}
    Masked Modeling (MM) has demonstrated widespread success in various vision challenges, by reconstructing masked visual patches.
%    aiming to reconstruct the masked patches, has demonstrated widespread success in various visual challenges. 
    Yet, applying MM for large-scale 3D scenes remains an open problem due to the data sparsity and scene complexity.
    % inherent spatial intricacy (\eg, complex furniture composition and structural layout).这种写法会把方法限制到室内场景
    % the unique complexity of the data. 
    % In this paper, we argue the inferior performance of such a self-supervised strategy on 3D scenes is constrained by the %inherent imbalanced redundancy. 
    % The standard random masking paradigm as used in 2D images often causes a high risk of uncertainty when reconstructing the underlying geometric structure.
    % The conventional random masking paradigm often causes a high risk of masking the basic geometric structure in a scene, making masked recovery extremely difficult.
    The conventional random masking paradigm used in 2D images often causes a high risk of ambiguity when recovering the masked region of 3D scenes.
    % as used in 2D images 
%    To this end, a novel guided masking strategy is proposed by exploring local statistics to preserve representative structured points, which  simplifies the pretext task. 
    To this end, 
    we propose a novel informative-preserved reconstruction,
    which explores local statistics to discover and preserve the representative structured points, effectively enhancing the pretext masking task for 3D scene understanding. 
    % and dropping massive redundant points.
    Integrated with a progressive reconstruction manner,
    % where a masked scene is restored into a relatively more complete one during each iteration, 
    our method can concentrate on modeling regional geometry and enjoy less ambiguity for masked reconstruction.
    %regional patterns with less ambiguity,
    % local geometry based on regional context
    % , instead of recovering the whole enormous space impetuously.
    % while overlooking inattentive reconstruction.
    % \mt{or: instead of recovering the whole enormous space inattentively}.
    % incremental, confound
    % 下面的这句话有点突兀，没有和前边的结合起来，
    % Further, a self-distillation strategy is introduced to extract the spatial consistency from visible areas, 
    % Besides, the target-input scenes during masked reconstruction serve as the natural teacher-student for self-distilling their intrinsic spatial consistency,% 不好理解
    Besides, 
    such scenes with progressive masking ratios can also serve to self-distill their intrinsic spatial consistency, 
    % \mt
    {requiring to learn the consistent representations from unmasked areas}.
    % 我不确定这么说，对吗，但是你们没有解释 consistency的作用。我感觉逻辑应该是，你们引入了啥，他作用是啥，最终形成了个啥。现在的写法，感觉consistency不是你的贡献，也不知道他用来干嘛
    % what is self-distillation masking-invariant
    By elegantly combining informative-preserved reconstruction on \textbf{masked} areas and consistency self-distillation from \textbf{unmasked} areas, a unified framework called \textbf{MM-3DScene} is yielded.
    % To verify the effectiveness of our method, 
    We conduct comprehensive experiments on a host of downstream tasks. The consistent improvement (\eg, +6.1\% mAP@0.5 on object detection and +2.2\% mIoU on semantic segmentation) demonstrates the superiority of our approach.
    % Project website: \url{ https://mingyexu.github.io/mm3dscene/}.
    %which explores rich local statistics to pick out representative XXX. 
    %\mt{straightforward?} guided masking strategy, which exploits XXX and restores XX.
    %\mt{which exploits local differences in terms of colors and coordinates to discover the points with latent dense semantics, and preserves them during masking. }
    %In doing so, it enjoys the benefit of XXX.
    %\mt{it encourages MM on 3D scenes (the self-supervised task) being neither ambiguous nor simple}
    %\mt{or: it avoids the pretext task being too confusing, while promises the high-level understandings of holistic scenes}
    %\mt{Based on this, we further propose a progressive masked reconstruction manner for the extensive understanding of 3D scenes and achieve decent performance based on masked modeling. 
    %Moreover, we introduce the momentum contrast that perfectly marries with our MM design, yielding a more powerful framework.}
    %Comprehenzsive experiments demonstrate the effectiveness of our method. XXX
    %Different from 2D images, where patches are dropped out randomly, 3D point cloud often exhibits 
    
\end{abstract}

\section{Introduction}
\label{sec:intro}

3D scene understanding plays an essential role in various visual applications, such as virtual reality, robot navigation, and autonomous driving. 
%Driven by the rise of 3D scene datasets \cite{sunrgbd,s3dis,scannet,toscene}, deep learning are dominantly applied in 3D scene parsing tasks \cite{pointtrans,votenet,pointgroup}. 
%, including 3D semantic segmentation \cite{paconv,pointtrans}, 3D object detection \cite{votenet,3detr}, and 3D instance segmentation \cite{pointgroup}.
Over the past few years, deep learning has dominated 3D scene parsing tasks \cite{pointtrans,votenet,pointgroup}.
However, traditional supervised learning methods require massive annotation of 3D scene data that are extremely laborious to obtain \cite{scannet}, where millions of points or mesh vertices per scene need to be labeled.

%Inspired by the success of contrastive-based SSL in 2D area \cite{lai2019contrastive,chen2020contrastive}, contrastive learning 

To solve this, self-supervised learning (SSL) becomes a favorable choice since it can extract rich representations without any annotation \cite{bert,moco,mae}.
% One convention of SSL, contrastive learning (CL)\cite{lai2019contrastive,chen2020contrastive}, has achieved impressive effectiveness in 3D scene understanding \cite{PointContrast,csc,4dcontrast}.
% One mainstream of SSL, masked modeling (MM) 
Masked Modeling (MM) \cite{mae,simmim},
as one of the representative methods in SSL, 
recently draws significant attention in the vision community.
Recently,
It has been explored in 3D vision \cite{OcCo,pointbert,pointmae,liu2022masked,zhang2022masked,zhang2022point},
where
% Borrowing the idea from MM on object-centric 2D images, 
% To define the pretext task for generating the self-supervision signals from the data themselves, 
these 3D MM approaches 
% directly inherit the idea from 2D, where they
randomly mask local regions of point clouds, and pre-train neural networks to reconstruct the masked areas.
% for single-object 3D shapes.
Nevertheless, such random masking paradigms are not feasible for large-scale 3D scenes, which often causes a high risk of reconstruction ambiguity.
% of masking the basic geometric structure of a scene.
As illustrated in Fig.~\ref{fig:intro} (a), a chair and a TV are totally masked, which are extremely difficult to be recovered without any context guidance.
% without guidance by any structural clues. 
Such ambiguity often makes MM difficult to learn informative representation for 3D scenes.
% In this case, large uncertainty are introduced, making it hard for neural networks to obtain useful representations.
%%%%%%
% As a result, they perform poorly when transferring into scene understanding tasks (see Table~\ref{tab:method_compare}).
% although they improve the baseline performance when fine-tuning on shape-level tasks, 
% In our pilot study, we also pre-trained Point-MAE on large-scale 3D scene datasets (\eg, ScanNet\cite{scannet}), but still got inferior results \mt{PointMAE in Table.~\ref{tab:ablations_mask_strategy})}.
%ignoring point-wise semantic segmentation task that is more challenging and predominant in 3D scene understanding, which hinders them to serve as a general SSL framework for 3D scene understanding. (only write voxel-mae in related work?)}
% Here we ask: \textit{what makes MM different between 3D scenes and shapes?}
Hence, we ask a natural question: \textit{can we customize a better way of masked modeling for 3D scene understanding?}

% Applying MM into large-scale 3D scenes remains an open problem.}

% Inspired by the analysis of He \etal \cite{mae} on 2D images, 
% We attempt to seek for a new solution by revisiting the data characteristic of 3D scenes.
% He \etal \cite{mae} first raised the concept of information density of image data, as the key philosophy behind MAE \cite{mae}. 
% They claim that while the human-generated languages hold very dense information (\eg, only a few missing words make the whole sentence be distinct from original meaning or be confusing), image signal is naturally redundant (\eg, a missing patch can be predicted from neighboring patches as a shortcut).
% As indicated by He \etal \cite{mae} and Pang \etal \cite{pointmae}, both image data and single-object 3D shape data have heavy information redundancy -- \eg, a missing patch can be easily predicted from neighboring patches (Fig.~\ref{fig:intro} (a)).
% , where various furniture is entangled with structural context under an intricate spatial layout (\eg, relative poses, directions, and distances). 
% Accordingly, as for mask strategy, language models often xxx, brutely masking makes xxx \cite{bert}.
% Following MM in 2D, a simple random mask with a very high ratio (80$\sim$90\%) is employed on 3D shapes to reduce the redundancy, and neural networks can still approximately infer the masked patches so that to gain high-level representations \cite{mae,simmim,pointmae,liu2022masked} \mt{(see Fig.~\ref{fig:information_density} (a))}.
% To begin with, we need to solve the potential problem caused by conventional random masking.
To tackle this question,
we propose a novel informative-preserved masked reconstruction scheme in this paper.
% ht: We remove the above ambiguity by concentrating on restoring local XXX. 
% To solve this potential problem, we propose an informative-preserved mask strategy.
% a Local Statistics Guided mask strategy.
Specifically, we leverage local statistics of each point (\ie, the difference between each point and its neighboring points in terms of color and shape) as guidance to discover the representative structured points which are usually located at the boundary regions in the 3D scene.
% with distinct information, which are hard to be parsed even under full supervision.
% Instead of being information-redundant,
We denote these points as `\textbf{Informative Points}' since they provide highly useful information hints and rich semantic context for \textit{assisting} masked reconstruction. 
%useful information and
%%% 这句写的虚了，没明白说什么
% , and a few masks on them may bring significant unpredictability.
To this end, our mask strategy is definite: to \textit{preserve} Informative Points in a scene and mask other points.
In this way, the basic geometric information of a scene is explicitly \textit{retained}, which effectively simplifies the pretext task and reduces ambiguity.
% Compared with conventional methods~\cite{pointbert,pointmae,OcCo,liu2022masked}, 
%% 我觉得写的不好，太highlevel，这句连上上局的解释，没有说清楚，informative point 你怎么用，为什么这样用，你就能改善masked modeling. 我看后面这一小段是描述怎么好的。前面这几句，我感觉得在总结总结，写的太虚。
% In short, the key philosophy behind our 3D mask strategy is \textit{{``a good pretext task is \textbf{not too difficult}''}}.

Based on our mask strategy, a progressive masked reconstruction manner is integrated, to better model the \textit{masked} areas.
As illustrated in Fig.~\ref{fig:intro} (b), 
during each iteration, 
% our method reconstructs a masked scene into a more \textit{relatively} complete one with a smaller masking ratio.
our method \textit{concentrates} on reconstructing the local \textit{regional} geometric patterns rather than rebuilding the original intact scene.
% , where models \textit{focus} on recovering \textit{regional} geometric .
% \textcolor{red}{
% (\eg, sometimes attends background surfaces, sometimes foreground instances),
% }
%这句话，没什么意义。
% , or skeletal contours) 
% instead of broadly wandering the whole scene with no concentration, 
% enjoying less ambiguity while promising to obtain not only diverse but also exact information, and yielding better results (Table~\ref{tab:ablation_progressive}).
In doing so, it enjoys less ambiguity and is able to restore accurate geometric information.

Moreover,
we realize the information of \textit{unmasked} areas (\ie, Informative Points) is underexplored.
%%%% 1 这段和前面两段的承接关系不好，不是说尽管成功，还有啥没做。你就直接说，还有啥没做，做了他怎么就好了
%%%% 2 这段前面的两段，mask什么，不mask什么，都没解释清楚。我知道你们的paper，我都感觉，到底mask什么东西？就这种感觉。所以前面两段要refine一下
We find that there exists point correspondence in the unmasked areas under progressive masking ratios.
% We find that the target-input scenes during masked reconstruction may serve as natural teacher-student, where the spatial consistency of Informative Points between target-input scenes can be \textit{self}-distilled.
%%%读不明白，不知道是个啥
% during the progressive masked reconstruction are natural self-supervised signals \textit{without} any extra rigid augmentations.
Accordingly, we introduce a dual-branch encoding scheme for learning such intrinsic consistency, with the ultimate goal of unearthing the consistent (\ie, masking-invariant) representations from unmasked areas.
This leads to a more powerful SSL framework on 3D scenes, called MM-3DScene,
which elegantly combines the masked modeling on the masked and unmasked areas in 3D scenes together, while \textit{complements} each other.
It achieves superior performance  in 
Table~\ref{tab:ablation_framework} $(\romannumeral 5)$.

Our contributions are motivated and comprehensive:

\begin{itemize}
% \vspace{-0.5em}
\item We raise the concept of Informative Points -- the points providing significant information hints, and indicate that preserving them is critical for assisting masked modeling on 3D scenes (Table \ref{tab:ablation_mask}).
% We investigate a critical masked modeling problem on 3D scenes -- A few 
% 你有实验验证吗？ 如果你这样做，就会结果大幅度的下降
% \vspace{-0.5em}
\item For masked areas, we propose an informative-preserved reconstruction scheme to focus on restoring the regional geometry in a novel progressive manner, which explicitly simplifies the pretext task.
% \vspace{-0.5em}
\item For unmasked areas, we introduce a self-distillation branch,
which is encouraged to learn spatial-consistent representations under progressive masking ratios.
% maintains the spatial consistency of unmasked points under different masking ratios.
% \vspace{-0.5em}
\item A unified self-supervised framework, called  MM-3DScene, delivers performance improvements across on various downstream tasks and datasets (Table~\ref{tab:performance_improvement}). 
% \vspace{-0.5em}
% \item A Local Statistics Guided mask strategy is introduced to preserve the Informative Points, which explicitly avoids the pretext task being too confusing.

% \item We propose to better recover masked areas in a novel progressive manner, and to self-distill the spatial consistency from unmasked areas, yielding a unified self-supervised framework for 3D scene understanding, called MM-3DScene.

% \item Extensive experiments on a number of widely used benchmarks demonstrate the effectiveness of MM-3DScene. \my{Performance gain:XX}
\end{itemize}

\begin{table}[t]
    \centering
    \resizebox{0.48\textwidth}{!}{
    \large
    \begin{tabular}{*{10}{l}}
        \toprule
       Datasets  &  Complexity & Task &  Gain (from scratch)  \\
        \midrule
        S3DIS & Entire floor, office & segmentation & (\textcolor{purple}{+1.5\%}) mIoU\\
        \midrule
         &    & segmentation & (\textcolor{purple}{+2.2\%}) mIoU\\
        ScanNet & Large rooms  & detection & (\textcolor{purple}{+4.4}) mAP@0.25\\ 
        &  & detection & (\textcolor{purple}{+6.1}) mAP@0.5\\ 
        \midrule
        SUN-RGBD & Cluttered rooms  & detection & {(\textcolor{purple}{+2.9}) mAP@0.25}\\ 
          &    & detection & {(\textcolor{purple}{+4.4}) mAP@0.5}\\ 
        \bottomrule
    \end{tabular}
    }
    \caption{\textbf{Summary of fine-tuning MM-3DScene} on various downstream tasks and datasets for 3D understanding. Our MM-3DScene conspicuously boosts the performance of the baseline model trained from scratch.}
    \vspace{-1.5em}
    \label{tab:performance_improvement}
\end{table}

% \begin{array}{llc}
% \hline & & \text { Sem. Seg. } \\
% \text { ScanNetV2} & 1.2 & \text { Ins. Seg. } \\
% & & \text { Obj. Det. }
% \end{array}

%-------------------------------------------------------------------------
\section{Related Work}
\label{sec:related_works}

\paragraph{3D Scene Understanding}
With the rapid development of deep learning methods in point cloud analysis and the emergence of large-scale 3D datasets \cite{scenenet,s3dis,scannet,Matterport3D,toscene},
the research focus gradually migrate from synthetic, single object analysis \cite{gsnet,gdanet,curvenet,spidercnn,rscnn,pointmlp} to complex large-scale scene understanding,
especially scene  segmentation \cite{pointweb,sparseconv, he_pointinst3d_2022,He2021dyco3d,He_memory_2022,xu2021investigate}, 3D object detection \cite{votenet,3detr,groupfree}.
% Previous approaches \cite{pointweb,} 
% They focus on exploring the point cloud representation and deep architectures for point cloud processing, which have made adequate progress on scene understanding tasks.
PointNet \cite{pointnet} and its variants \cite{pointnet2,dgcnn,pointcnn,pointconv} extract local features from neighbors through hierarchical grouping architecture to capture fine-grained representation.
Thomas \etal \cite{kpconv} defines deformable kernel point convolution to capture the point cloud representation using a set of learnable kernel points.
In PAConv \cite{paconv}, a continuous convolutional kernel is built by dynamically combining several weight banks, where the coefficients are learned from point positions.
Zhao \etal \cite{pointtrans} proposes to assign learnable attentional weights to local point features, and introduces a Transformer \cite{transformer}-like architecture into point cloud analysis.
% which is proved to be an effective backbone for various scene-understanding tasks.
Rather than focusing on developing deep architectural details, 
in this paper, we explore an effective self-supervised pre-training mechanism for scene understanding  based on the basic version of Point Transformer.
% use the basic version of Point Transformer as our backbone to explore effective pre-training mechanisms

\paragraph{Unsupervised Scene Pre-training}
Self-supervised learning (SSL) has recently achieved great success in 2D vision \cite{chen2020simsiam,byol,moco,chen2021mocov3,bao2022beit,mae,simmim} and NLP tasks \cite{gpt,bert,gao2021simcse}. 
% Due to the complex pre-training tasks and large-scale model parameters, 
% Self-supervised pre-training models can implicitly encode large-scale data and store it in the parameters of the model, which in turn enables downstream task gains. 
% Compared to 2D vision and NLP, 
But there has been limited exploration of SSL for 3D vision.
Most of the existing 3D-SSL methods aim to understand 3D point clouds, and can be divided into two types, masked modeling (MM)-based methods, and contrastive learning-based methods, respectively.

MM-based pre-training usually takes as input point clouds to recover itself as the pretext task.
OcCo \cite{OcCo} first constructs an occlusion point cloud and applies an encoder-decoder mechanism to reconstruct the original object.
Yu \etal \cite{pointbert} proposes the idea of restoring the masked proxies under the supervision of the pre-trained tokenizer.
Borrowing the idea from MAE \cite{mae}, Pang \etal \cite{pointmae} designs a masked auto-encoder to recover the masked parts of objects.
However, most of these MM-based methods are focused on 3D shape-level pre-training,
how to apply such a scheme on 3D scenes has not been fully investigated.
% considering that point cloud scenes are much more complex and noisy than objects. 
Although some concurrent works \cite{voxmae,Voxel-MAE} voxelize the outdoor automative point clouds and randomly mask voxels, they are only manifested to especially benefit the LiDAR-based object detection task. 
% Thus, a generic MM-based SSL framework for 3D scene understanding is still lacking.
In this paper, we creatively apply the self-supervised pre-training directly on 3D scenes by a scene-specific masked reconstruction.
% mechanism.

On the other hand, using the idea of contrastive scene contexts for pre-training has been shown to be effective \cite{PointContrast,csc,4dcontrast}.
Xie \etal \cite{PointContrast} and Hou \etal \cite{csc} perform scene pre-training using contrastive learning of point features between a pair of overlapping scans.
4DContrast \cite{4dcontrast} first composites the synthetic 3D object with real-world 3D scans to create 4D sequential data, 
and utilizes the contrastive loss to learn 4D invariance constraints.
For our MM-based framework, we leverage the momentum contrast \cite{moco} to self-distill the consistency between the same scenes under different masking degrees, for unearthing the hidden information from unmasked areas.

\section{Method}%
\label{sec:method}

\subsection{Overview}

\begin{figure*}[tb] \centering
    \includegraphics[width=0.99\textwidth]{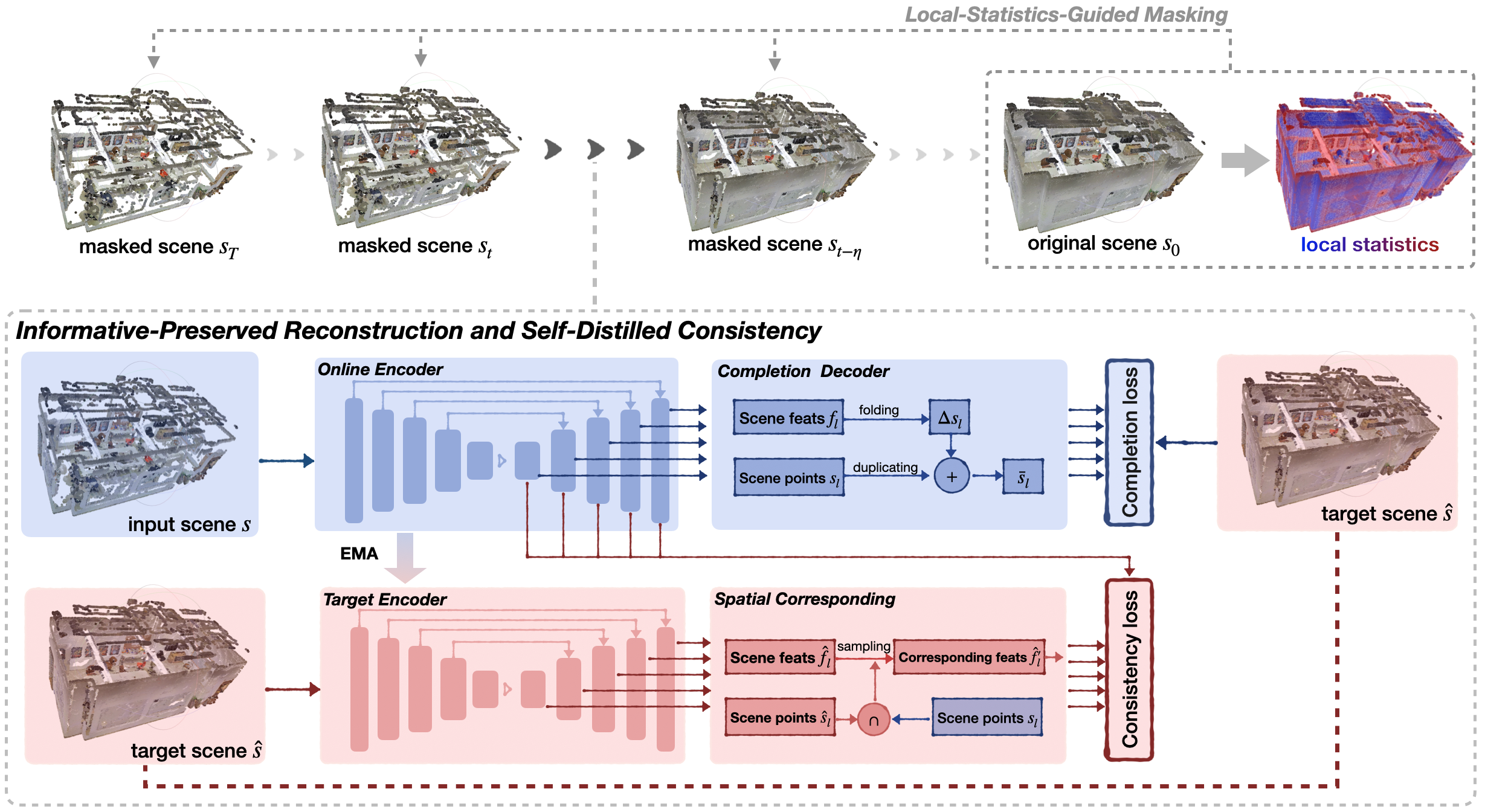}
    \caption{Overview of our customized masked modeling framework \textbf{MM-3DScene}. 
    We first propose the Local-Statistics-Guided masking strategy to discover and  preserve the representative structured points.
    % which provides useful information clues to assist the masked reconstruction. 
    This mask strategy yields an Informative-Preserved Reconstruction, where our method focuses on restoring the \textit{regional} geometric patterns of \textbf{masked areas} at each learning step,
    enhancing the pretext masking task with less ambiguity.
    Moreover, since \textbf{unmasked areas} are underexplored during reconstruction, 
    we introduce a self-distillation branch to maintain the intrinsic \textit{spatial consistency} under progressive masking ratios, which enables MM-3DScene to learn consistent (\ie, masking-invariant) representations of the unmasked areas.
    % the model is encouraged to maintain the intrinsic \textit{spatial consistency} on unmasked points between different masking ratios, which requires the \mt{masking-invariant} understanding of unmasked regions.
    } \label{fig:method_overview}
    \vspace{-1em}
\end{figure*}

As shown in Fig.~\ref{fig:method_overview}, we propose MM-3DScene for masked modeling on 3D scenes,
which can effectively alleviate the uncertainty and unpredictability during masked reconstruction for enhancing the self-supervised pre-training on complex 3D scenes.
 
Firstly, we propose a Local-Statistics-Guided masking strategy in Sec.~\ref{sub:mask_strategy}, which explores the local difference for discovering and preserving Informative Points during the scene masking.
Next, our mask strategy is integrated with a progressive reconstruction manner, yielding an informative-preserved reconstruction scheme.
% where our method focuses on restoring local regional geometry.so as to enjoy less ambiguity and is able to restore accurate geometric information. 
% to explore the representations from unmasked regions, 
Finally, we introduce a self-distillation branch for learning the spatial consistency in the unmasked areas under progressive masking ratios. 
By elegantly combining informative-preserved reconstruction and self-distilled consistency, MM-3DScene is yielded.
% can boost performance on various downstream tasks. 
The framework details are presented in Sec.~\ref{sub:network}.

\begin{figure}[tb] \centering
    \includegraphics[width=0.39\textwidth]{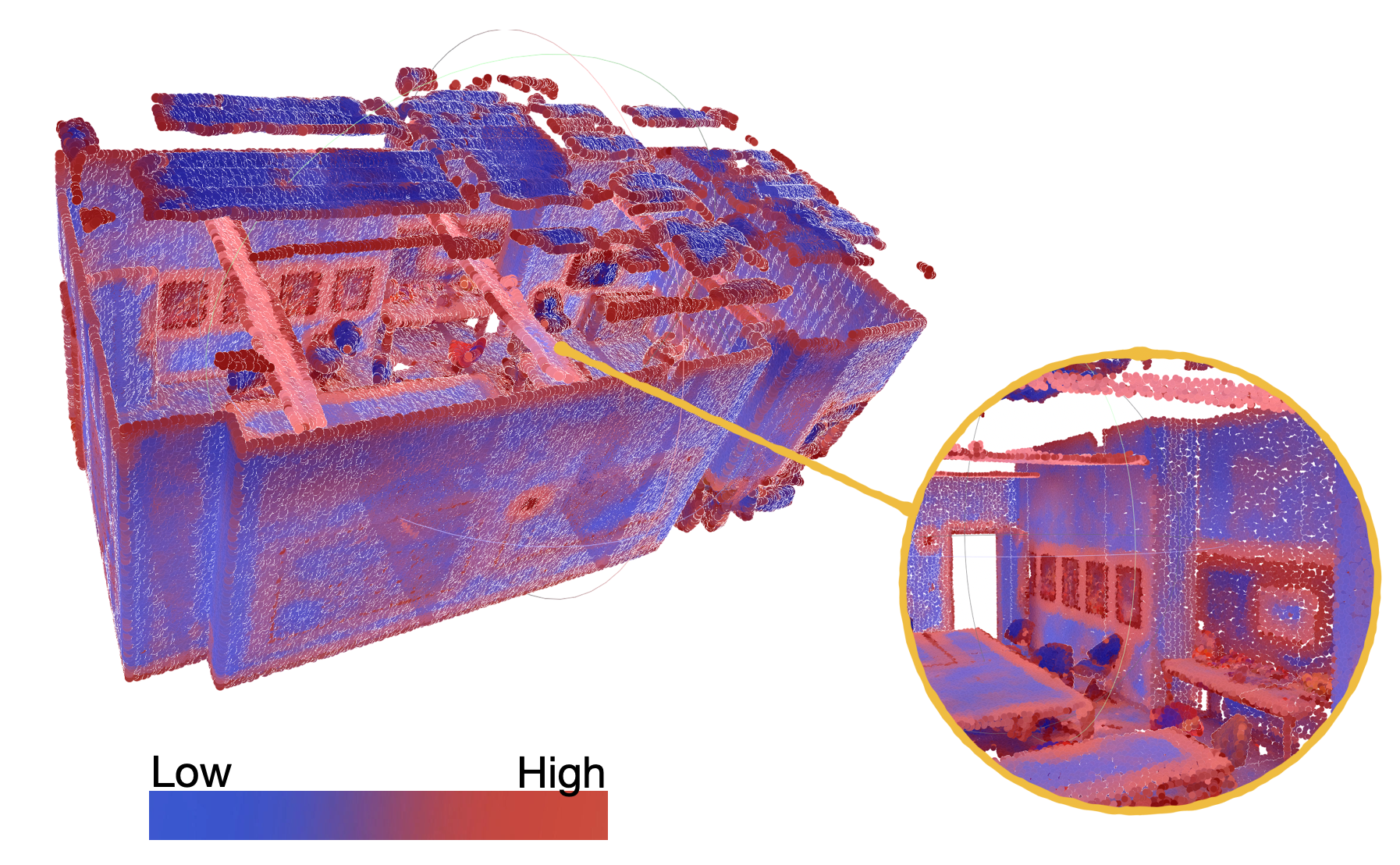}
    \caption{\textbf{The heat map of local statistics on 3D scenes}. Points with high statistics value also provide highly important semantic information for understanding or reconstructing 3D scenes.} \label{fig:method_LDvis}
\end{figure}

% masked scene is reconstructed into a to model the meaningful geometry patterns,
% as illustrated in Sec. \ref{sub:mask_strategy}.
% Specifically, for the input masked scene, we choose another masked scene that is a bit more complete than it as the completion target, instead of completing the input directly to the original scene.
% Via such design, 
% the completion uncertainty can be weakened,
% while the model can focus on recovering the perceptible geometry patterns.
% we can effectively weaken the uncertainty of scene completion and 
% \my{increase the diversity of pre-trained pairs.}
% We will introduce the details of our mask modeling manner in Sec. \ref{sub:mask_strategy}.

% which is shown in Figure \ref{fig:method_overview}.
% In addition to using a progressive scene completion as a pretext task, we also learn the spatial consistency of the dense points at different mask scales.
% 
% then bring the spatial corresponding feature from two branch closer.
% find the corresponding point features in the target scene

% informative-preserved mask modeling and spatial consistency learning at different mask scales,
% our pre-training framework can further boost the performance on various downstream tasks. 

% use the idea of comparative learning

% In our MM3DScene pre-training, the input scene is $S_t$, which is randomly selected from the mask scene sequence $\mathcal{S}$, and the corresponding completion target is $S_t-1$. 
% As shown in  Figure \ref{fig:method_overview}, 
% % we extract scene features using the current more effective feature extraction backbone, and then use the scene features for two pre-training tasks.
% our pre-training task consists of two parts: progressive  complemention of scenes and spatial correspondence representation learning.

% This section introduces our proposed MM-DScene for Pre-training. 
% % Please organize your method into several parts.
% First we introduce the scene-specific mask strategy,
% and then we elaborate our MM3DScene Mask Modeling  framework based on our mask strategy.

% \subsection{Overview}%
% \label{sub:Overview}

\subsection{Local-Statistics-Guided Masking}%
\label{sub:mask_strategy}

As we analyzed in Sec.~\ref{sec:intro}, the traditional random mask strategy is not feasible for complex large-scale 3D scenes, which often causes a high risk of ambiguity during masked reconstruction.
% Considering that many foreground objects in the scene may be masked out in their entirety.
% % and  such mask strategy breaks the background structural  context  of the scene.
% Thus the scene under such mask strategy may be difficult to be recovered,
%even under full supervision, 
% causing large uncertainty in self-supervised model learning.
In order to explore a more effective masked modeling mechanism for 3D scenes,
% we need to develop a new mask strategy for the complex scenes.
we need to first design a better mask strategy, aiming to \textit{reduce the ambiguity} during the masked reconstruction pretext task.

In 3D scenes, some representative structured points in provide highly important information hints and rich semantic context for assisting the scene understanding or reconstruction tasks.
Accordingly, preserving them may help a lot to simplify the reconstruction.
Thus, the first question is: \textit{how to find such points?}

% develop a new mask strategy by preserving some informative points during the scene masking.
% In this way, the pre-trained model can leverage the prior knowledge (e.g., parts of the foreground objects or structures of the background surface) to guide completion in the 3D scenes. 
% \noindent
% \textbf{Local statistics on scenes.}~
% First of all, 
% % we need to measure the local statistics of each point in the scene.
% we need to find a way of measuring the local statistics in the scene.
% Inspired from  IAFNet \cite{xu2021investigate}
In this paper, we adopt local statistics as straightforward guidance to discover the representative points.
For complex 3D scenes,
we use the local difference of each point (\ie, the difference between each point and its neighboring points in terms of colors and coordinates) to calculate the local statistics of each point.
% guidance to discover Informative Points.
% To a certain extent, 
% local statistics can be represented by the local difference between each point and its neighbors, which is inspired from  IAFNet \cite{}.
% \mt{here I suggest directly copying the description in introduction.}
% For example, the local geometry changes gently on the surface areas of the wall, which shows lower information density.
% While the geometric structures are complex and varied on the boundary areas between wall and window, and the information density of such areas is relatively higher.
% \mt{we should directly say points with redundant information and points with dense information, which is redundant points and informative points, as indicated in abstract and introduction.}
% So we use local geometric difference as a criterion of measuring the information density.
% \mt{as the guidance to discover informative areas/points, avoid mention information density, it is hard to be `measured'.},
Specifically,
for each point $p_i$ in the scene, 
we first use K-Nearest Neighborhood (KNN) search to obtain its $K$ neighboring points $p_{ik}$ in Euclidean space,
and then calculate the local difference to denote point statistics:
% (i.e., coordinates, colors and fine-grained features):
% \begin{equation}
%     D_{xyz}(p_i) =  \sum_{k=1}^K\left\|p_{i,xyz}-p_{i_k,xyz}\right\|_2
% \end{equation}
% \begin{equation}
%     D_{rgb}(p_i) =  \sum_{k=1}^K\left\|p_{i,rgb}-p_{i_k,rgb}\right\|_2
% \end{equation}
% \begin{equation}
%     D_{emb}(p_i) =  \sum_{k=1}^K\left\|p_{i,emb}-p_{i_k,emb}\right\|_2
% \end{equation}
\begin{equation}
    D_{q}(p_i) =  \sum_{k=1}^K\left\|p_{i,q}-p_{i_k,q}\right\|_2
\end{equation}
where $p_{i,q} \in \mathrm{R}^{1 \times C_q}$  is point statistic of  $p_i$.
Specifically,
$p_{i,0}$ and $p_{i,1}$ are the point coordinates and colors,  
% while $p_{i,2}$ indicates the feature embedding from  a point-based embedding block (e.g., PointNet++ \cite{pointnet2} or PointTransformer block \cite{pointtrans}).
% and the point-wise feature  $p_{i,emb} \in \mathrm{R}^{C}$ is extracted from the first point embedding layer (e.g., PointNet++ \cite{} or PointTransformer block \cite{}).
% Then we normalize and accumulate these differences together as the information density $D \in R^{N\times 1}$, where $N$ is the number of the points in the scene.
which are normalized and accumulated together as the local statistics.
% \begin{equation}
$
    D = \sum_{q=0}^{1} (\alpha_q \times \textit{norm}( D_{q})) 
$.
% \end{equation}
where $D \in \mathrm{R}^{N\times 1}$, $N$ is the number of the points in the scene.

% \my{
For a clearer picture, we visualize the heat map of local statistics on the 3D scene in Fig.~\ref{fig:method_LDvis}, 
it can be seen that points with high statistics value
%% 这个词不对，应该是值很高吧，value？？ 或其他。
are concentrated in the foreground objects or contours of the scene (\textit{red regions}), 
and these areas are relatively more informative and provide highly important information hints for understanding or reconstructing a 3D scene, denoted by \textbf{`Informative Points'} in this paper.
% and these regions can be combined to characterize the structural information of the scene.
% To conclude, we use the local statistics to discover the Informative Points in 3D scenes.
% \textbf{Local-statistic-guided mask strategy.}
Informative Points provide highly useful
information hints and rich semantic context for \textit{assisting} masked reconstruction.
To this end, our mask strategy is definite: to \textit{preserve} Informative Points in a scene and mask other points.
Through this, the basic geometric information
of a scene is explicitly \textit{retained}, which effectively reduces ambiguity in the pretext masking task.

% In order to preserve the regions with informative points over scene masking.
% For pre-training, we randomly select a masked scene $S_{t}$ form scene sequence $ \mathcal{S} $ as the input scene.
% \my{To further weaken the uncertainty of the model for scene completion, and XXX} 
% we choose scene $S_{t-1}$ as the complete target for input scene $S_{t}$. 
% Meanwhile, compared to complete the original scene $S_0$ from $S_t$,  such training pairs can effectively enhance the data diversity for pre-training,
% which is verified in Sec. \my{XX}.

% find its K nearest neighbor points in Euclidean space using KNN search method, and then calculate the local difference value of its nearest neighbor region
% the wall area, the geometric structure changes gently, the information density is relatively low, while the boundary area of beams or windows, the geometric structure changes in a complex and diverse way, the information density of these areas is relatively higher

% \textcolor{lightgray}{\lipsum[1]}

\subsection{MM-3DScene}%
\label{sub:network}
% \mt{Refer to Fig. 3, and change the contrast loss in the figure into Consistency Loss}
After introducing the mask strategy with local statistics guidance,
we illustrate the details of our MM-3DScene -- a self-supervised masked modeling framework customized for 3D scene understanding.
Our pre-training architecture consists of two parts: informative-preserved reconstruction and self-distillation of spatial consistency, respectively aiming at better modeling of masked areas and unmasked areas in 3D scenes.
These two parts are elegantly combined while \textit{complementing} each other in our MM-3DScene.
%% 空间一致行吗？空间老感觉是2维。是不是应该写3D consistency。我在前面改了一出。你们自己斟酌一下。 
% On the one hand,
% we feed the input masked scene into the online encoder to extract visual representation,
% and  then utilize the completion decoder to generate complementary points of the masked areas.
% % on the masked scene.
% On the other hand,
% to learn the masking-invariant representation of unmasked areas,
% we design a concise dual branch self distillation strategy for self-distill the intrinsic spatial consistency.
% Specifically, 
% we introduce a target encoder to deal with target scene,
% then narrow the point feature of the two branches for which spatial correspondence exists.
% In our MM3DScene pre-training, the input scene is $S_t$, which is randomly selected from the mask scene sequence $\mathcal{S}$, and the corresponding completion target is $S_t-1$. 
% As shown in  Figure \ref{fig:method_overview}, 
% % we extract scene features using the current more effective feature extraction backbone, and then use the scene features for two pre-training tasks.

% \subsubsection{Progressive Completion Manner}
% \subsection{Point Transformer backbone}%
% \textcolor{lightgray}{\lipsum[1]}
% \noindent\textbf{Progressive Completion Manner.}
\paragraph{Informative-preserved reconstruction.}
Based on the proposed mask strategy, we introduce an informative-preserved reconstruction manner.
% This stage is designed to recover the local regional geometric patterns, based on informative masked scenes with progressive masking ratios.
% \textcolor{red}{a great amount of } %% 怪怪的，是说数量多，是吗？
% Here we introduce our mask strategy for the generating the pre-training pairs.

To begin with, for the original scene $s_0$, 
we align all scene points in descending order according to the local statistics $D$. 
Then we use the \textit{incremental} masking ratio $\mathcal{\theta} = \{ \theta_1, ..., \theta_t,...,\theta_T \}$ to \textit{progressively} mask the scene $s_0$ according to $D$ from low to high, forming the scene sequence $ \mathcal{S} = \{s_1,...s_{t-1}, s_t,...,s_T\}$ as shown in the Fig. \ref{fig:method_overview}.
It can be seen that in the masked sequence $\mathcal{S}$,  the masked regions are gradually shifted from the background surface to the foreground objects,
and the representative structural points are preserved.
It is also worth noting that scene $s_t$ is the \textit{subset} of scene $s_{t-1}$.

During each training iteration,
we randomly select a masked scene $s_t$ in sequence {$\mathcal{S}$} as the input,
and take a \textit{relatively} more complete one $s_{t-\eta}$ as the target for masked reconstruction, 
% whereT he masking ratio of input scene $s_t$ is $\theta_t$,
% while the masking ratio of target $s_{t-\eta}$ is $\theta_{t-\eta}$, 
where $\eta$ indicates the masked gap to be recovered (to facilitate the subsequent description, we later define the input scene as $s=s_t$ and the target as $\hat{s}=s_{t-\eta}$).
To encode point-wise representations from the input scene $s$, 
we utilize a well-established network $\Phi_{OE}$  as our backbone.
$\Phi_{OE}$ is a point based feature extractor with hierarchical structure.
The hierarchical  encoded features $\mathcal{F}$ of scene $s$ can be represented as
% \begin{equation}
   $ \mathcal{F} = \Phi_{OE}(s) $
% \end{equation}
where $\mathcal{F} = \{ f_1, ...,f_l,...,f_L\}$,
and $L$ is the layers of $\Phi_{OE}$.
% $f_l \in \mathrm{R}^{N_l \times C}$,

% To obtain  more useful representation, 
% we utilize all the hierarchical features of $\mathcal{F}$ to complete the scene $s$  at each sample  layer.
Then we utilize the features extracted from the online encoder to learn the coordinate variations for scene reconstruction.
Note that the number of output points $\hat{s}$ is larger than that of input points $s$.
% \my{
Before generating coordinate variations via MLP,
we follow SnowFlakeNet\cite{xiang2021snowflakenet}
% \textcolor{red}{[reference]} 
to replicate the coordinates and features,
producing more points for  outputs.
% }
Specifically,
for $l$-th layer, we have the displacement feature $f_l$ and scene points $s_l$.
% We generate the coordinate variations via a FoldingNet \cite{} style module.
Referring to FoldingNet \cite{foldingnet},
we incorporate a standard two-dimensional grid $I$ to the displacement feature $f_l$, 
and then use two consecutive Multi-Layer Perceptrons (MLPs) \cite{mlp} to generate coordinate variation $\Delta s_l$: 
\begin{equation}
    \Delta s_l=\Psi_2(f_l \oplus \Psi_1(f_l \oplus I)),
\end{equation}
where $\Psi_1$ and $\Psi_2$ are two 3-layer MLPs, 
and $\oplus$ is the concatenation operator.
Then, we add  the coordinate variation $\Delta s_l$ with the duplicated input mask scene $s$,
which generates the predicted scene $\bar{s}_l$ in the $l$-th layer.
Finally, in order to train the masked reconstruction task more efficiently,
we choose the multi-scale symmetrical chamfer distance \cite{topnet} as the reconstruction loss,
with the following details:
\begin{equation}
    \mathcal{L}_{PC}
    = \sum_{l}^{L} (
    \frac{1}{|\bar{s}_l|} \sum_{x \in \bar{s}_l} \min _{y \in \hat{s}}\|x-y\|^2+\frac{1}{|\hat{s}|} \sum_{y \in \hat{s}} \min _{x \in \bar{s}_l}\|x-y\|^2).
\end{equation}

The proposed reconstruction scheme encourages the model to \textit{focus} on restoring the \textit{regional} geometry in a novel progressive manner, which enjoys less ambiguity and is able to restore accurate geometric information, so to enhance the modeling on masked areas.

% complement the scenes separately using all the hierarchical features.

\paragraph{Consistency self-distillation.}
Moreover, we realize the information of unmasked areas (\ie, Informative Points) is underexplored. 
We find that there exists point correspondence in the unmasked areas under progressive masking ratios.
Leveraging this unique behaviour of our method, and 
based on a generalized model of teacher-student mutual learning \cite{zhang2019your,gou2021knowledge},
we introduce a self-distillation branch to maintain the intrinsic spatial consistency on unmasked areas during progressive reconstruction.

% To generate more stable representation of the unmasked areas,
% we use a self-distilling learning model to enhance the consistency between the input  and target scene.
% It is 
% but is uniquely designed for progressive scene modeling.
% % \my
% {
% considering that more accurate local geometric representation can be extracted from the  target scene.}
% \textcolor{red}{but is uniquely designed for progressive scene modeling.}
%%% 我感觉你没写出特色，他会就觉得这就是MOCO之类的东西。你这章，主要在写你怎么做。你要想突出特色，你应该写，你为什么这么做。
As Fig. \ref{fig:method_overview} shows,
we treat the online encoder as the student model and maintain a teacher model as the target encoder.
It is worth mentioning that the target encoder has the same parameters and structure as the online encoder,
but does not participate in the back-propagation of the gradient.
The parameters of the target encoder are dynamically updated by the  Exponential Moving Average (EMA) \cite{gou2021knowledge}:
\begin{equation}
    \mathcal{W}_{\text {target}} \leftarrow [ \beta \mathcal{W}_{\text {target }}+(1-\beta) \mathcal{W}_{online}],
\end{equation}
where $\mathcal{W}_{\text {online}}$ and $\mathcal{W}_{\text {target}}$ are the parameters of online encoder and target encoder.

Next, we feed the target scene $\hat{s}$ to the target encoder and extract the target feature $\hat{\mathcal{F}} = \{ \hat{f}_1, ...,\hat{f}_l,...,\hat{f}_L\}$, where $\hat{f}_l \in \mathrm{R}^{\hat{N}_l \times C}$.
Considering the input scene is the subset of target scene, where
$s \subset \hat{s}$,
we can select a subset of the target feature $\hat{f}_l $ that have natural spatial correspondence with the online feature ${f}_l$.
We define such subset of target feature as $\hat{f}_l^{\prime} \in \mathrm{R}^{N_l \times C} $,
where $N_l < \hat{N}_l$.
% Inspired by PointContrast \cite{},
Finally, we use info-NCE loss \cite{oord2018representation} to model the spatial consistency of the feature representation between the input scene and the target scene, which can be formulated as:
% The loss for self-distilling consistency 
\begin{equation}
    \mathcal{L}_{CSD}=-\sum_{l}^{L} \sum_{(i, j) \in s_l} \log \frac{\exp \left({f}_{i,l} \cdot \hat{f}_{j,l}^{\prime}/ \tau\right)}{\sum_{(\cdot, k) \in s_l} \exp \left({f}_{i,l} \cdot \hat{f}_{k,l}^{\prime} / \tau\right)},
\end{equation}
where the input scene $s_l$ is the set of points that can find the spatial correspondences in the target scene $\hat{s}_l$.
For online  feature ${f}_{i,l}$ of point $p_i$, 
we take the corresponding point feature $\hat{f}_{j,l}^{\prime}$ in the target scene as the positive sample,
and use feature $\hat{f}_{k,l}^{\prime}$ as the negative samples, 
where $\exists(\cdot, k) \in s_l$ and $k \neq j$.
% Inspired by PointContrast \cite{},
% we random set 4096 sampled pairs from each scene.

Through self-distillation, our method is able to unearth the spatial-consistent (\ie, masking-invariant) representations from unmasked areas. 

\paragraph{MM-3DScene.}
Ultimately, by elegantly combining informative-
preserved reconstruction on masked areas and consistency self-distillation from unmasked areas, a unified framework called MM-3DScene is yielded.
The final pre-training loss can be denoted by $\mathcal{L} = \zeta_1 \times \mathcal{L}_{PC} +  \zeta_2 \times \mathcal{L}_{CSD}$,
We find that setting both $\zeta_1$ and $\zeta_2$ to 1 yields the best result, which means that the masked reconstruction and the consistency distillation share \textit{balanced importance} for our framework.

% \subsubsection{Backbone Method}

% \begin{table*}[t] \centering
%     \newcommand{\Frst}[1]{\textcolor{red}{\textbf{#1}}}
%     \newcommand{\Scnd}[1]{\textcolor{blue}{\textbf{#1}}}
%     \caption{Please put the caption text in the main tex file, and put the table code to ``table/xxx.tex''. It will be easier for others to modify. A two-column table with two sub-tables. \Frst{Red} text indicates the best and \Scnd{blue} text indicates the second best result, respectively.}
%     \label{tab:table2}
%     \input{tables/two_subtables}
% \end{table*}

\section{Experiments}%
\label{sec:Experimental Setup}
We pre-train our MM-3DScene and demonstrate its effectiveness on a variety of generic downstream tasks for 3D scene understanding, including 3D semantic segmentation, 3D object detection, as well as data-efficient setup and linear probing evaluation.
% \mt{ADD result improvement in all tables!!!}

\subsection{Experimental Setup}
\label{sub:setup}

\paragraph{Datasets.}~Following the state-of-the-art 3D self-supervised pre-training methods \cite{PointContrast,csc,4dcontrast}, we pre-train the proposed MM-3DScene on ScanNetv2 \cite{scannet} dataset with 1201 train scans.
As for 3D object detection, we fine-tune the pre-trained weights on ScanNetv2 training set (inner-domain) and evaluate on the corresponding validation set with 312 scenes.
We also fine-tune our model on SUN RGB-D \cite{sunrgbd} training set for cross-domain evaluation.
SUN RGB-D covers fully 3D object bounding box annotations for 10 categories, with 5,285 train frames and 5,050 test frames.
As for downstream semantic segmentation, 
in addition to conducting the  experiments on the ScanNetv2,
we also evaluate our pre-trained models on S3DIS \cite{s3dis}, 
which contains 271 rooms from 3 different buildings.
% another indoor scene dataset 

% As for 3D semantic segmentation, we fine-tune the pre-trained weights on ScanNetv2 training set (inner-domain) and evaluate on the corresponding validation set with 312 scenes.
% We also fine-tune our model on S3DIS \cite{s3dis} training set for cross-domain evaluation.
% S3DIS is a large-scale indoor scene dataset containing 271 rooms from 3 different buildings.
% % \mt{to be completed on S3DIS \cite{s3dis}}
% Similarly, we use ScanNetv2 for in-domain verification on 3D object detection tasks, and SUN RGB-D \cite{sunrgbd} for cross-domain transfer. SUN RGB-D covers fully 3D object bounding box annotations for 10 categories, which is split into 5,285 train frames and 5,050 test frames.

\begin{table}[tb]\centering
    \resizebox{0.48\textwidth}{!}{
    \large
    \begin{tabular}{{l} *{10}{c}}
        \toprule
       Method & \text{ } &  mAP@0.25 & mAP@0.5  \\
        \midrule
        DSS \cite{DSS}  & \text{ } &  15.2 & 6.8 \\
        F-PointNet \cite{qi2017frustum}   &\text{ } &  19.8 & 10.8 \\
        GSPN \cite{yi2018gspn} &\text{ } &  30.6 & 17.7 \\
        3D-SIS \cite{hou2019sis} & \text{ } & 40.2 & 22.5 \\
        % \midrule
        % PointContrast \cite{PointContrast} (scratch) & \text{ } & 55.6 & 31.7\\
        % PointContrast \cite{PointContrast}  & \text{ } & 57.5 & 34.8\\
        \midrule
        VoteNet \cite{votenet} (scratch) & \text{ } &  58.7 & 35.4 \\
        RandomRooms \cite{RandomRooms} + VoteNet & \text{ } &  61.3 \small\textcolor{gray}{(+2.6)}  & 36.2 \small\textcolor{gray}{(+0.8)} \\
        PointContrast \cite{PointContrast} + VoteNet & \text{ } & 58.5 \small\textcolor{gray}{(-0.2)}  & 38.0 \small\textcolor{gray}{(+2.6)} \\
        CSC \cite{csc} + VoteNet & \text{ } & - & 39.3 \small\textcolor{gray}{(+3.9)} \\
        4DContrast \cite{4dcontrast} + VoteNet  & \text{ } &  - & 40.0 \small\textcolor{gray}{(+4.6)} \\
        \textbf{MM-3DScene (w/o $L_{CSD}$) + VoteNet}  & \text{ } &  \textbf{61.9 \small\textcolor{purple}{(+3.2)} } & \textbf{41.3 \small\textcolor{purple}{(+5.9)}}\\
        \textbf{MM-3DScene + VoteNet} & \text{ } &  \textbf{63.1 \small\textcolor{purple}{(+4.4)}} & \textbf{41.5 \small\textcolor{purple}{(+6.1)}}\\
        % \midrule
        % \my{H3DNet\cite{h3dnet}} & \text{ } & \my{64.8} & \my{47.4}\\
        % \my{\textbf{MM-3DScene + H3DNet}} & \text{ } & \my{\textbf{66.8 \small\textcolor{purple}{(+2.0)}}} & \my{\textbf{48.9 \small\textcolor{purple}{(+1.5)}}} \\
        \bottomrule
    \end{tabular}
    }
    \caption{3D object detection results on ScanNetv2.}
    \label{tab:scannet_detection}
\end{table}

\begin{table}[tb]\centering
    \resizebox{0.48\textwidth}{!}{
    \large
    \begin{tabular}{{l} *{5}{c}}
        \toprule
       Method & \text{ } & mAP@0.25  & mAP@0.5  \\
       \midrule
       DSS \cite{DSS}  & \text{ } &  42.1 & - \\
       COG \cite{COG} & \text{ } &  47.6 & - \\
       2D-driven \cite{2Ddriven} & \text{ } &  45.1 & - \\
       F-PointNet \cite{qi2017frustum}  & \text{ } &  54.0  & - \\
        \midrule
        VoteNet \cite{votenet} (scratch) & \text{ } & 57.7  & 32.9 \\
        PointContrast \cite{PointContrast} + VoteNet  & \text{ } & 57.5 \small\textcolor{gray}{(-0.2)}   & 34.8 \small\textcolor{gray}{(+1.9)} \\ 
        RandomRooms \cite{RandomRooms} + VoteNet & \text{ } &  59.2 \small\textcolor{gray}{(+1.5)}  & 35.4 \small\textcolor{gray}{(+2.5)} \\
        % DeepContrast \cite{deepcon} + VoteNet & \text{ } &  61.6 & 35.5\\
        CSC \cite{csc} + VoteNet & \text{ } &  -  &  36.4 \small\textcolor{gray}{(+3.5)} \\
        4DContrast \cite{4dcontrast} + VoteNet & \text{ } & - & 38.2 \small\textcolor{gray}{(+5.3)}  \\
        % \textbf{MM-3DScene (w/o $L_{CSD}$) + VoteNet}  &   & 59.5 & 35.9\\
        \textbf{MM-3DScene + VoteNet} & \text{ } & \textbf{60.6 \small\textcolor{purple}{(+2.9)}}  &\textbf{37.3 \small\textcolor{purple}{(+4.4)}}\\
        \bottomrule
    \end{tabular}
    }
    \caption{3D object detection results on SUN RGB-D.}
    \label{tab:exp_sunrgbd}
\end{table}

\begin{figure}[tb] 
\centering
    \includegraphics[width=0.37\textwidth]{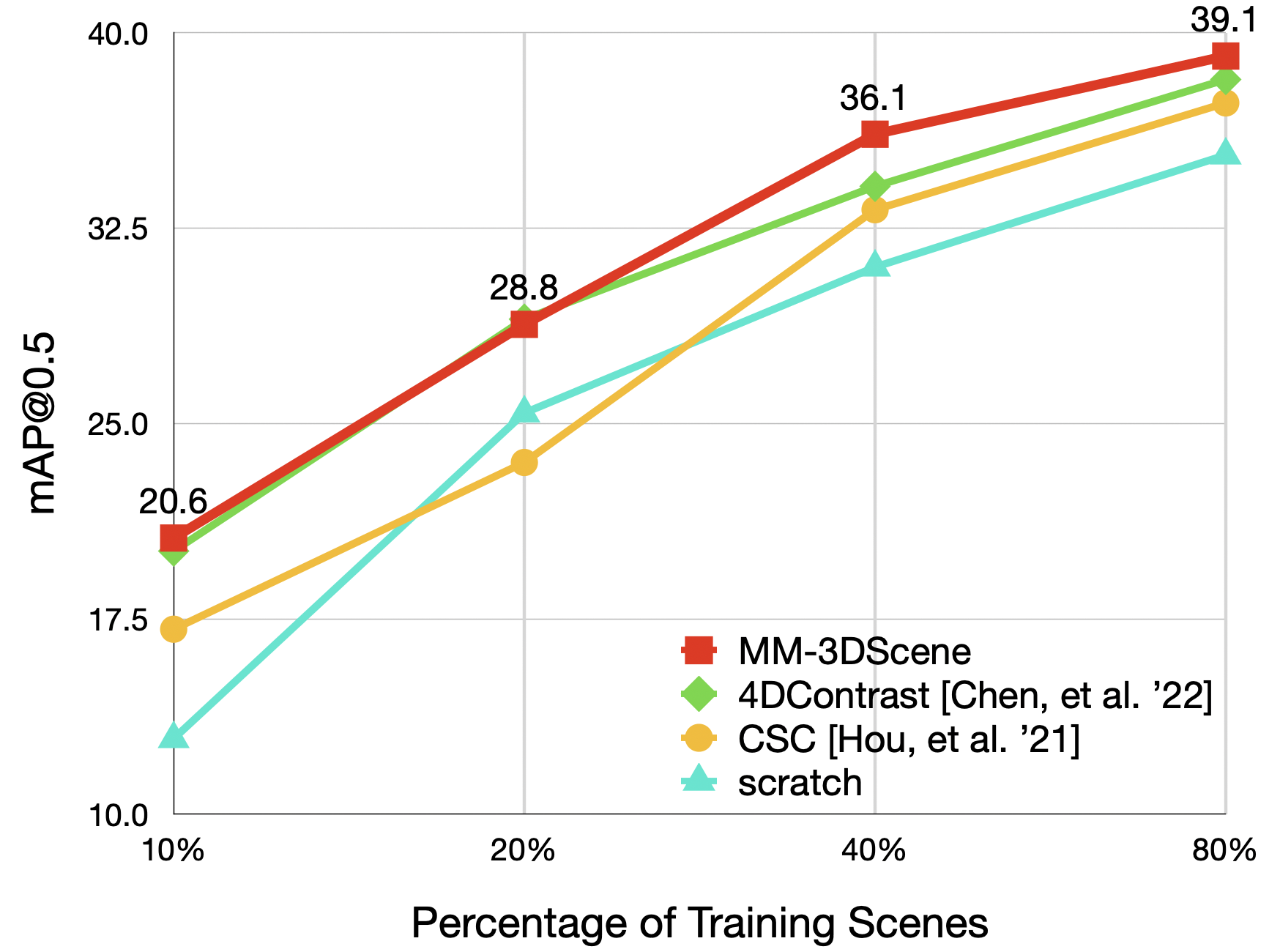}
    \caption{Data-efficient 3D object detection results on ScanNetv2.}
    \label{fig:exp_data_efficent}
    \vspace{-1em}
\end{figure}

\begin{table}[tb]\centering
    \resizebox{0.48\textwidth}{!}{
    \large
    \begin{tabular}{{l} *{5}{c}}
        \toprule
       Method   & \text{mAcc}   & mIoU  \\
        \midrule
        PointNet++ \cite{pointnet2}  & \text{-} & 53.5 \\
        PointConv \cite{pointconv}   & \text{-}& 61.0  \\
        PointASNL \cite{yan2020pointasnl}   & \text{-}& 63.5 \\
        FPConv \cite{lin2020fpconv}&   \text{-}& 64.4 \\
        KPConv \cite{kpconv}  & \text{-}& 69.2\\
        % SparseConvNet \cite{} &\text{    }& \text{ } & 69.3\\
        % PointTransformer \cite{pointtrans} & \text{-} & 70.6 \\
        % MinkowskiNet \cite{} &  & - & 65.4 \\
        \midrule
        SR-UNet \cite{mink}   & 78.1 & 70.0\\ 
        CSC\cite{csc} + SR-UNet    &   78.8 \small\textcolor{gray}{(+0.7)}  & 70.7 \small\textcolor{gray}{(+0.7)}\\
        PointContrast\cite{PointContrast} + SR-UNet   & 79.3 \small\textcolor{gray}{(+1.2)} & 71.3 \small\textcolor{gray}{(+1.3)} \\
        4DContrast \cite{4dcontrast} + SR-UNet   & 80.8 \small\textcolor{gray}{(+2.7)} & 72.3  \small\textcolor{gray}{(+2.3)}\\
        \midrule
        PointTrans \cite{pointtrans} (scratch)     & 79.6 & 70.6  \\
        % CSC \cite{csc} + PointTransformer &\text{    }& \text{ } & XX \\
        PointMAE \cite{pointmae} * + PointTrans    & \text{79.6 \small\textcolor{gray}{(+0.0)} } & 70.6 \small\textcolor{gray}{(+0.0)}\\
        PointContrast \cite{PointContrast} * + PointTrans   & \text{80.0 \small\textcolor{gray}{(+0.4)} }& 70.9 \small\textcolor{gray}{(+0.3)}\\
        % \midrule
        % scratch & \text{    }& \text{ } & 69.0 \\
        \textbf{MM-3DScene (w/o $L_{CSD}$) + PointTrans}  & \textbf{81.2 \small\textcolor{purple}{(+1.6)}}& \textbf{{72.1 \small\textcolor{purple}{(+1.5)}}}\\
        \textbf{MM-3DScene + PointTrans}  & \textbf{82.0 \small\textcolor{purple}{(+2.4)}}& \textbf{{72.8 \small\textcolor{purple}{(+2.2)}}}\\
        \bottomrule
    \end{tabular}
    }
    \caption{3D semantic segmentation results on ScanNetv2.}
    \label{tab:exp_scannet_seg}
\end{table}

\begin{table}[tb]\centering
    \resizebox{0.48\textwidth}{!}{
    \small
    \begin{tabular}{{l} *{10}{c}}
        \toprule
       Method &  mAcc & mIoU  \\
        \midrule
        PointNet \cite{pointnet} & 49.0 & 41.1 \\
        PointCNN \cite{pointcnn} & 63.9 & 57.3 \\
        PointWeb \cite{pointweb} & 66.6 & 60.3 \\
        % MinkowskiNet \cite{mink} & - & 71.7 & 65.4 \\
        PAConv \cite{paconv}  & 73.0 & 66.6 \\
        KPConv \cite{kpconv} & 72.8 & 67.1 \\
        % PointTransformer \cite{pointtrans} & 90.8 & 76.5 & 70.4\\
        % StratifiedTransformer \cite{lai2022stratified} & 91.5 & 78.1 & 72.0\\
        % \midrule
        % Scratch + SR-UNet \cite{mink}& - & 75.5 & 68.2\\
        % PointContrast \cite{PointContrast} + SR-UNet  & - & 77.0 & 70.9\\
        % CSC \cite{csc} + SR-UNet  & - & - & 72.2\\
        \midrule
        % {Scratch} & {90.3} & {75.6} & {69.8}\\
        {PointTrans \cite{pointtrans}} (scratch)  & 76.5  & 70.4\\
        PointMAE  \cite{pointmae} * + PointTrans & 76.4 \small\textcolor{gray}{(-0.1)}  &  70.4 \small\textcolor{gray}{(+0.0)} \\
        PointContrast \cite{PointContrast} * + PointTrans  & 76.9 \small\textcolor{gray}{(+0.4)} & 70.7 \small\textcolor{gray}{(+0.3)}\\
        % \textbf{MM-3DScene (w/o $L_{CSD}$) + PointTrans}  & \textbf{77.2 \small\textcolor{purple}{(+0.7)}} & \textbf{71.1 \small\textcolor{purple}{(+0.7)}}\\
        \textbf{MM-3DScene + PointTrans}  & \textbf{78.0 \small\textcolor{purple}{(+1.5)}} & \textbf{71.9 \small\textcolor{purple}{(+1.5)}}\\
        \bottomrule
    \end{tabular}
    }
    \caption{3D semantic segmentation results on S3DIS Area-5.}
    \vspace{-1em}
    \label{tab:exp_s3dis}
\end{table}

\paragraph{Backbone networks.}~
% \mt{Instead of using 2D Transformer in MAE and Point-MAE who project the local point patches into token embeddings, which may hurt point-wise fine-grained understanding. We advocate modeling at the point level and use point-based backbone networks. Also easily to perform our masked strategy.}
% \mt{to be completed for ScanNet pre-training on Point Transformer and S3DIS segmentation...}
In 3D object detection, we follow \cite{PointContrast,csc,4dcontrast} to use VoteNet \cite{votenet} as the backbone. 
% \textit{Notably}, VoteNet originally applies PointNet++ \cite{pointnet2} for feature extraction, but the aforementioned approaches all replace the original PointNet++ layers with SR-UNet \cite{sparseconv,mink} and pre-train on SR-UNet. In contrast, 
We \textit{directly} perform MM-3DScene pre-training on the original PointNet++ layers, for minimizing the backbone effect on performance gains and better verifying the stand-alone effectiveness of pre-trained representations.
Besides, we follow the data augmentations in the original backbone networks.
For the scene semantic segmentation, we advocate using Point Transformer \cite{pointtrans} as our backbone, considering its lightweight and effective attributes for 3D scene understanding (Table \ref{tab:exp_modelsize}).
We retain its original network architectures,
% the five-stage hierarchical architecture with the down-sampling rates of [1,4,4,4,4].
% We use the transition block \cite{pointtrans} to connect features of consecutive stages.
and discard the output head in the pre-training process in order to obtain the fine-grained feature representation.
The output head will be added to the network structure for the downstream training.

\paragraph{Implementation details.}~
% \mt{to be completed for ScanNet pre-training on Point Transformer and S3DIS segmentation...}
For pre-training, we use AdamW \cite{adamW} optimizer with a weight decay of 0.0005.
The initial learning rate is set to 0.001 and decayed at the 60\% and 80\% epochs.
We pre-train the network for 300 epochs with a batch size of 8.
The reconstruction gap $\eta$ is set to 0.1.
 We set the scale parameter  $\tau$  as 1.

For downstream semantic segmentation,
we follow the setting of Point Transformer \cite{pointtrans},
and utilize SGD optimizer with a momentum of 0.9 and weight decay of 0.0001.
In addition to the standard data augmentation, 
we also use random cropping of the scene for more effective training.
For S3DIS, we fine-tune our model for 150 epochs with a data loop of 30,
where the voxel size is 4cm.
The data loop for ScanNet is set to 6 and the scene resolution is 2cm.

For downstream object detection, the fine-tuning settings all follow 4DContrast \cite{4dcontrast}, where the networks are trained for 500 epochs, the initial learning rate is set to 0.001 and decayed by a factor of 0.5 at epoch 250, 350, 450.  
The batch size is 6 for ScanNet and 16 for SUN RGB-D.

% \begin{figure}[tb] 
% \centering
%     \includegraphics[width=0.5\textwidth]{imgs/fig_seg_vis.png}     \caption{Qualitative results on S3DIS semantic segmentation. } 
%     \label{fig:vis}
% \end{figure}

\subsection{3D Object Detection}
\label{sub:obj_det}

\paragraph{ScanNetv2 object detection.}
3D object detection is a widely used downstream task for scene understanding.
We report the fine-tuning results of object detection for ScanNetv2 in the Table \ref{tab:scannet_detection},
our MM-3DScene achieves the state-of-the-art performance among the SSL methods. 
With of MM-3DScene pre-training, 
it  achieves a significant improvement of \textbf{4.4} on mAP@0.25,
and \textbf{6.1} on mAP@0.5
compared to training from scratch.
Meanwhile,
even if we  train without the consistency distillation, 
the result also surpasses the state-of-the-art method 4DContrast \cite{4dcontrast},
which strongly demonstrates the effectiveness of our customized masked modeling method for object detection.
% \my{
Note that,
\textit{without} RGB, 
our method still gets \textbf{41.5} mAP@0.5 on ScanNetv2 detection,
which is much higher than 4DC (40.0) and PC (38.0).
Moreover, we conduct experiments with the updated H3DNet\cite{h3dnet} as the backbone and achieve significant improvements, which can be found in appendix.
% }

\paragraph{SUN RGB-D object detection.}
We also conduct the cross-domain experiments on SUN RGB-D dataset for object detection,
which is shown in Table \ref{tab:exp_sunrgbd}.
% \my{}
We pre-train our method on ScanNetv2 and fine-tune on SUN RGB-D,
for which our method substantially exceeds the  training-from-scratch baseline 
and surpasses  3D-based pre-training of  CSC \cite{csc}, PointContrast \cite{PointContrast} and RandomRooms \cite{RandomRooms}.
\textit{Notably}, pretraining 4DContrast \cite{4dcontrast} requires a \textit{prerequisite} non-trivial generation of spatio-temporal correspondence, while our method produces the masked scenes \textit{on the fly}.

% Alongside point-wise understanding, it is also important to parse scenes at the instance level.

\paragraph{Data-efficient evaluation.}
% \label{sub:data_effi}
As shown in Fig. \ref{fig:exp_data_efficent}, 
We evaluate  our method for fine-tuning with limited training data,
where the results of CSC and 4DContrast  are officially borrowed from \cite{4dcontrast}.
Our MM-3DScene surpasses both of them in almost all data-efficient settings.
It is worth noting that our method achieves \textbf{39.1} mAP@0.5 when finetuning with only \textbf{80}\% training data, 
which even exceeds   many methods trained on 100\% of the training data in Table \ref{tab:scannet_detection}.

\begin{figure}[tb] 
\centering
    \includegraphics[width=0.45\textwidth]{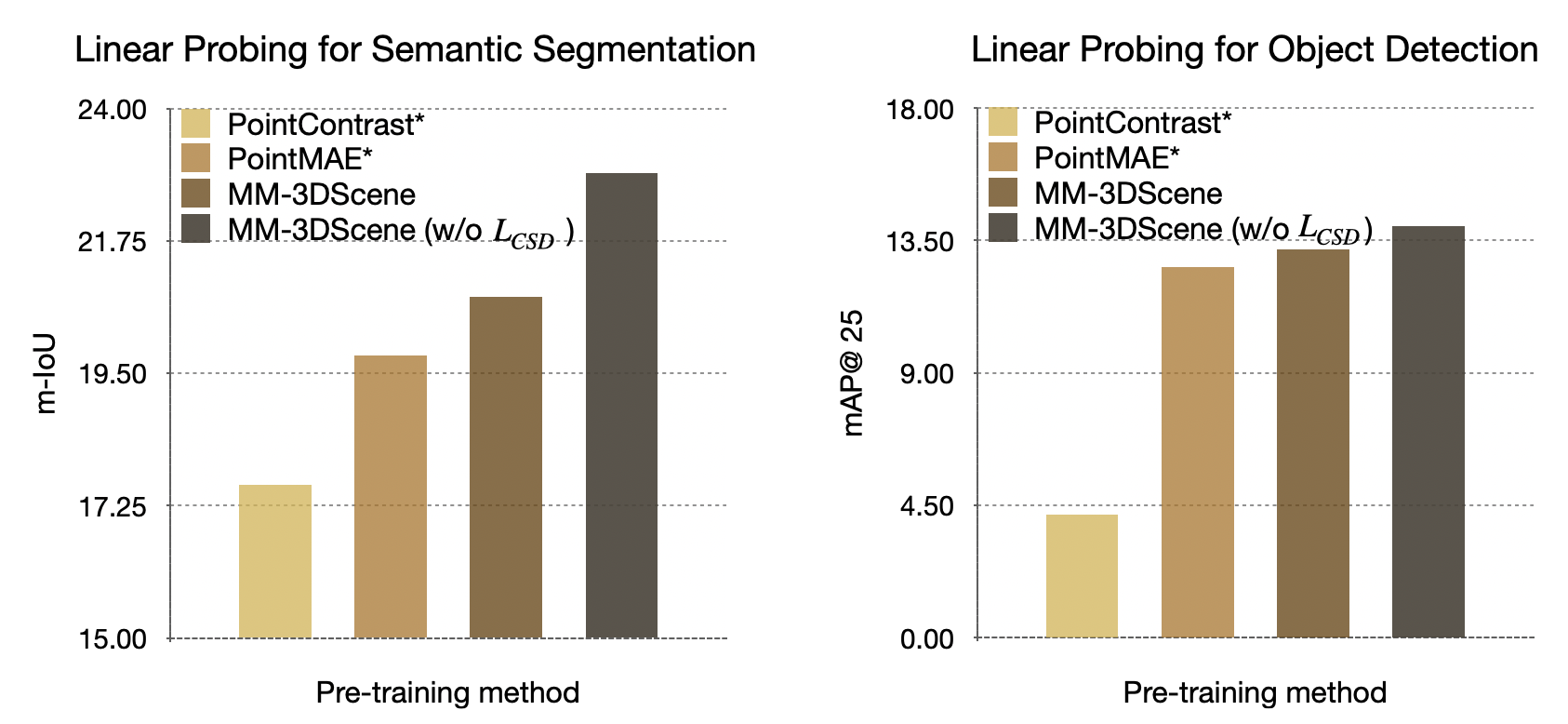} 
    \caption{Linear probing on  semantic segmentation and  object detection. } 
    \label{fig:exp_linearProbing}
    % \vspace{-1em}
\end{figure}

\subsection{3D Semantic Segmentation}
\label{sub:sem_seg}

\noindent
\textbf{ScanNetv2 semantic segmentation.}~
Semantic segmentation is a common but challenging task. It requires models to predict the semantic classes of each point, which is involved in the fine-grained understanding of 3D scenes.
Table \ref{tab:exp_scannet_seg} lists the downstream semantic segmentation results on ScanNetv2.
% \my{
For SR-UNet based methods, we directly quote the results from \cite{4dcontrast}, which do not use the RGB information as input.
However, for Point Transformer backbone, the RGB information is indispensable and omitting it will cause a deteriorated performance drop.
Thus, we follow its original input setting (points + rgb) and apply various pre-training methods on it.
We denote the methods with * when they are adapted to Point Transformer backbone, e.g., ``PointMAE\cite{pointmae}* + PointTrans". 
The table shows that the random masking strategy of PointMAE does not help the downstream segmentation task. 
We also observe that PointContrast* \cite{PointContrast} has a marginal improvement over the scratch model, which may indicate that contrastive learning is good at learning discriminative features but not robust to occlusions (as suggested by \cite{cma,msc}). 
In contrast, our MM-3DScene achieves superior results over these conventional pre-training methods. 
Furthermore, our MM-3DScene also outperforms 4DContrast with SRUNet backbone, while having a much lighter model size (Table~\ref{tab:exp_modelsize}).
% }
% \com{remove the rgb column from  table 4?}
% \DIFdel
% {
% Point-MAE \cite{pointmae} is a representative of conventional random masking.
% For a fair comparison, we apply the random mask modeling strategy of PointMAE to the same backbone as ours, defined as ``PointMAE* + PointTrans".
% The result  demonstrates that the uncertainty brought by such random masking does not benefit the downstream segmentation.
% Meanwhile, we also verify that PointContrast \cite{PointContrast} has a relatively small enhancement effect on the scratch,
% suspecting that contrastive learning only excels at learning discriminative features while overlooking the occlusions (as in \cite{cma,msc}). 
% By comparison, our MM-3DScene outperforms these representative conventional pre-training methods.  
% In addition, our MM-3DScene also beats other state-of-the-art pre-training methods that use SR-UNet as their backbone (heavy network architecture as in Table \ref{tab:exp_modelsize}), such as CSC \cite{csc} and 4DContrast \cite{4dcontrast}.
% Moreover, we stand out with a conspicuously more lightweight model than 4DContrast (Table~\ref{tab:exp_modelsize}).
% }

\noindent
\textbf{S3DIS semantic segmentation.}~
To further validate the effectiveness of our method, we conduct  experiments on the S3DIS dataset \cite{s3dis},
the results are shown in Table \ref{tab:exp_s3dis}.
We achieve a competitive result of 71.9\% mIoU on S3DIS Area-5,
which gets a relative improvement of 1.5\% from scratch.
% and the result  demonstrates that the uncertainty brought by such random masking does not bring  performance gains  to the downstream tasks.
% that random mask modeling .
% By comparison, our MM-3DScene performs better than other SSL methods, achieving a relative improvement of 1.5\%  m-IoU on S3DIS and 2.2\% on ScanNetv2.
% Note that even without spatial consistency learning, our method can still get a nicer performance than other self-supervised methods.
The qualitative results are illustrated in appendix.

\subsection{Linear Probing}
Linear probing is \textit{underexplored} in 3D scene understanding,
we remedy this defect by locking the pre-trained backbone network,
and only fine-tuning the segmentation head and detection head.
The experiments are conducted on S3DIS Area-5 and the ScanNetv2 validation set.
Fig. \ref{fig:exp_linearProbing} shows
results following different pre-training methods. 
Compared to the conventional random masking method (\ie, PointMAE \cite{pointmae}) or contrastive-based method (\ie, PointContrast \cite{PointContrast}), our MM-3DScene performs best on the linear probing task.
It is worth noting that our framework without learning spatial consistency performs better on linear probing,
we guess the reason is that \textit{fewer constraints} can enhance the generalizability.

% \mt{} 

\begin{table}[tb]\centering
    \resizebox{0.48\textwidth}{!}{
    \small
    \begin{tabular}{{l} *{5}{c}}
        \toprule
       Method  & & Model Size &\text{ }  & mIoU \\
        \midrule
        SR-UNet \cite{mink} (scratch) &\text{ } & 37.85M  &\text{ } & 70.0 \\ 
        4DContrast (3D) \cite{4dcontrast} + SR-UNet &\text{ } & 70.85M  \small\textcolor{gray}{(+33M)} &\text{ }  & 71.7 \small\textcolor{gray}{(+1.7)}  \\
        4DContrast (4D) \cite{4dcontrast} + SR-UNet  &\text{ } & 75.85M  \small\textcolor{gray}{(+38M)} &\text{ } & 72.3  \small\textcolor{gray}{(+2.3)}  \\
        \midrule
        PointTrans \cite{pointtrans} (scratch) &\text{ } & 7.76M &\text{ }  & 70.6 \\
        MM3DScene + PointTrans &\text{ } & 8.63M \small\textcolor{purple}{(+0.87M)} &\text{ }  & 72.8 \small\textcolor{purple}{(+2.2)}\\
        % MM3DScene + PointTrans  & 16.39M \small\textcolor{purple}{(+8.63M)}\\
        \bottomrule
    \end{tabular}
    }
    \caption{Comparisons of model parameters for different pre-training methods.}
    \vspace{-1em}
    \label{tab:exp_modelsize}
\end{table}

% \begin{table}[tb]\centering
%     \caption{Generalization Test}
%     \label{tab:table1}
%     \resizebox{0.43\textwidth}{!}{
%     \scriptsize
%     % \setlength\tabcolsep{19 pt}
%     \begin{tabular}{{l} *{5}{c}}
%         \toprule
%         Pretraining Model & m-IoU  \\
%         \midrule
%         PointMAE \cite{pointmae} + PointTransformer &  19.8  (val)\\
%         PointContrast \cite{PointContrast} + PointTransformer & 17.6(val) \\
%         MM-3DScene + Point Transformer & 22.9 (val)\\
%         \bottomrule
%     \end{tabular}
%     }
% \end{table}

\subsection{Ablation Studies}
\label{sub:ablation}

\noindent
\textbf{Why MM-3DScene is a better way for masked modeling on large-scale scenes?}~
Table \ref{tab:ablation_framework} shows the ablation study of our MM-3DScene framework on S3DIS semantic segmentation.
When we only use the informative masked modeling,
 ($\romannumeral 2$) achieves an improvement of 0.7\% mIoU over the train-from-scratch method.
On the other hand,
focusing on the unmasked areas (setting $\romannumeral 3$),
we only learn the spatial consistency of the unmasked informative points during each iteration, 
which still makes performance gains for downstream segmentation.
Finally, 
setting $\romannumeral 4$ integrates informative-preserved modeling of the masked areas and 
 spatial consistency learning on unmasked areas together,
which further boosts the result to 71.9\% mIoU.

% For the mask-off areas, 
% it is difficult for the network to learn useful representations with this kind of modelling approach with great uncertainty.
% On the other hand,
% focusing on the unmasked areas (setting $\romannumeral 3$),
% we only learn the spatial consistency of the unmasked informative points during each iteration, 
% which still make performance gains for downstream segmentation.
% Finally, 
% setting $\romannumeral 4$ marries the masked modeling on the masked and unmasked areas in 3D scenes together,
% which further boost  the downstream performance.

\noindent
\textbf{How MM-3DScene simplifies the pretext task?}~
Table \ref{tab:ablation_mask} shows the ablation study on mask strategies.
Comparing the settings of (a) and (d) in Table \ref{tab:ablation_mask},
we found that our informative-preserved masked modeling can significantly improve the downstream  performance, 
% of 3D scene downstream task, 
while random mask modeling does not.
The possible reason is that such masking will randomly drop Informative Points, thus bringing great uncertainty to the pretext task.
With the same consistency loss, PointMAE gets 71.0\%, which is 0.9\% lower than ours.
While our approach can make the masked areas perceptible via informative-preserved masking.
Instead, when we use the \textit{informative-abandoned} masking setting (c) by masking off the informative structured points,
the downstream performance is significantly reduced,
demonstrating that masking the Informative Points may cause large unpredictability.
% Setting (b) and (e)  demonstrate  that progressive manner can improve the effectiveness of both random masking and informative-preserved masking.

Moreover, different from (d),
setting (e) utilizes the progressive manner by reconstructing a masked scene into a more complete one,
and shows a significant improvement.
We also apply the progressive reconstruction manner based on the random masking (setting (b)), and the results are improved accordingly.

% when we directly complete the mask scene to the original complete scene, 
% our MM-3DScene achieves a slight performance boost compared to random masked modeling,
% by reason of preserving the informative points.
% However, there  still remains some ambiguity in the pre-training model, as  direct recovering of the  original complete scene from the unmasked scene is still difficult and unperceptible.
% Thus,  the pretext task is not so difficult and we reduce the ambiguity of masked modeling for complex scenes.
% Moreover, we also apply the progressive completion manner on the random masked modeling (setting (b)), and the results are improved accordingly.

% However, the large complementary span will still cause confusion to the model, 
% although the informative points are preserved, the large complementary span will still cause confusion to the model, 
% and the result is reflected in the Table \ref{tab:ablation_pretrain_pairs} (a),

% \mt{progressive steps: 3 ways comparison}
% \mt{Use feature difference in local statistics.} Supple

\noindent
\textbf{Memory cost.}~
Current 3D pre-training methods \cite{PointContrast,csc,4dcontrast}
% (\eg, PointContrast \cite{PointContrast}, CSC \cite{csc} and 4DContrast \cite{4dcontrast})  
basically adopts SR-UNet \cite{mink} as the backbone with 37.85M parameters,
We advocate the much more lightweight Point Transformer \cite{pointtrans}, with just 7.76M parameters.
In Table \ref{tab:exp_modelsize}, we compare our model size with 4DContrast \cite{4dcontrast} which uses SR-UNet.
Our MM-3DScene \textit{only adds} 0.87M parameters to the backbone.
% while performing better on ScanNet semantic segmentation.
Besides, our approach does not require the time-consuming pre-generation of contrastive point cloud pairs for pretraining used in PointContrast \cite{PointContrast} and CSC \cite{csc}.
Our masked scenes are generated \textit{on the fly} during network training.

\begin{table}[tb]\centering
    \resizebox{0.47\textwidth}{!}{
    \large
    \begin{tabular}{{c}|{l} {l} *{5}{c}}
        \toprule
      & Pre-training method & Pre-training loss  & mIoU & mAcc  \\
        \midrule
      $\romannumeral 1$ & \textcolor{gray}{Scratch} & -  & \textcolor{gray}{70.4} & \textcolor{gray}{76.5} \\
       % $\romannumeral 2$ & Random Mask as \cite{pointmae}  & $\mathcal{L}_{PC}$ 
       % & 70.4 \small\textcolor{gray}{(+0.0)}  & 76.4 \small\textcolor{gray}{(-0.1)}  \\
      $\romannumeral 2$ &  MM-3DScene (MM only) &   $\mathcal{L}_{PC}$ 
       & 71.1  \small\textcolor{gray}{(+0.7)} & 77.2  \small\textcolor{gray}{(+0.7)}\\
      $\romannumeral 3$ &  MM-3DScene (consistency only) &   $\mathcal{L}_{CSD}$ 
       &  70.9   \small\textcolor{gray}{(+0.5)}& 77.1   \small\textcolor{gray}{(+0.6)}\\
      $\romannumeral 4$ &  MM-3DScene  &  $\mathcal{L}_{PC}$, $\mathcal{L}_{CSD}$ 
       &  71.9  \small\textcolor{purple}{(+1.5)} & 78.0  \small\textcolor{purple}{(+1.5)}\\
        \bottomrule
    \end{tabular}
    }
    \caption{{Ablation study on MM-3DScene framework on S3DIS semantic segmentation.}}
    \label{tab:ablation_framework}
\end{table}
% \mat

\begin{table}[tb]\centering
    \resizebox{0.47\textwidth}{!}{
    \large
    \begin{tabular}{{c}|{l} {l} *{5}{c}}
        \toprule
       & Masked Modeling & Mask Strategy & Progressive   & mIoU & mAcc  \\
        \midrule
        &  \textcolor{gray}{Scratch} & -  & - & \textcolor{gray}{70.4} & \textcolor{gray}{76.5} \\
      (a) & PointMAE* \cite{pointmae} & Random  & $\times$ & 70.4 & 76.4 \\
      (b) & PointMAE* \cite{pointmae} & Random  & $\checkmark$ & 70.6 & 76.5 \\
      (c) & MM-3DScene (MM only) & Informative-abandoned   & $\times$ & 70.2 & 76.1\\
      % (d) & MM-3DScene (MM only) & Informative-abandoned   & \times  & 70.3 & 76.4\\
      (d) & MM-3DScene (MM only) & Informative-preserved  & $\times$ & 70.9 & 76.9 \\
       % (d) & MM-3D Scene (MM only) & \checkmark  & 71.3 & 77.3\\
      (e) & MM-3DScene (MM only)  & Informative-preserved  & $\checkmark$ & 71.1 &  77.2\\
      (f) & MM-3DScene  & Informative-preserved  & $\checkmark$  & 71.9 &  78.0\\  
        \bottomrule
    \end{tabular}
    }
    \caption{Ablation study on  mask strategies on S3DIS semantic segmentation.}
    \vspace{-1em}
    \label{tab:ablation_mask}
\end{table}

\section{Conclusion}%
\label{sec:Conclusion}
We have presented MM-3DScene, a customized masked modeling framework for 3D scene understanding. It explicitly preserves the representative structured points, which provides highly useful information clues to simplify the pretext task of masked reconstruction.
At each learning step, a masked scene is reconstructed in a progressive manner, so that to focus on restoring regional geometry and enjoy less ambiguity. 
% encourages the model to
Moreover, a self-distillation branch is integrated for maintaining the intrinsic spatial consistency on unmasked areas under the progressive masking ratios.
Extensive experiments on various downstream tasks verified 
% the effectiveness of our  MM-3DScene.
that our MM-3DScene significantly boosts the performance of baseline models trained from scratch.

\paragraph{Limitations.}
% The Informative Points in our approach are discovered by analyzing local statistics, which is an explicit way to find and preserve the representative points, effectively providing useful information hints to simplify the pretext task.
% We suspect there possibly exist other elegant ways to precisely define and locate the areas with the richest information, we leave this for future explorations.
In this paper, we mainly focus on indoor scenes, following recent self-supervised methods \cite{csc,4dcontrast} for 3D scene understanding.
We believe that the generic design insight of our masked modeling may inspire more researchers to solve 3D outdoor perception, 3D shape understanding, and 2D image recognition.

\section*{Acknowledgments}
This work was supported in part by the National Key R\&D Program of China (NO.~2022ZD0160505),  and in part by  the Youth Innovation Promotion Association of Chinese Academy of Sciences (No.~2020355).
It was also partially supported by NSFC62172348, Outstanding Youth Fund of Guangdong Province with No.~2023B1515020055 and Shenzhen General Project with No.~JCYJ20220530143604010, the National Key R\&D Program of China with grant No.~2018YFB1800800, by Shenzhen Outstanding Talents Training Fund 202002, by Guangdong Research Projects No.~2017ZT07X152 and No.~2019CX01X104, by the Guangdong Provincial Key Laboratory of Future Networks of Intelligence (Grant No.~2022B1212010001), and by Shenzhen Key Laboratory of Big Data and Artificial Intelligence (Grant No.~ZDSYS201707251409055).
Besides, thanks to Ji Hou for helpful suggestions on experiments.

\clearpage
%%%%%%%%% REFERENCES
{\small
\bibliographystyle{ieee_fullname}
\bibliography{ref}
}

% \clearpage
% \section*{Appendix}

\section*{Appendix A: Visualization Results}
\begin{figure}[!htbp] 
\centering
    \includegraphics[width=0.5\textwidth]{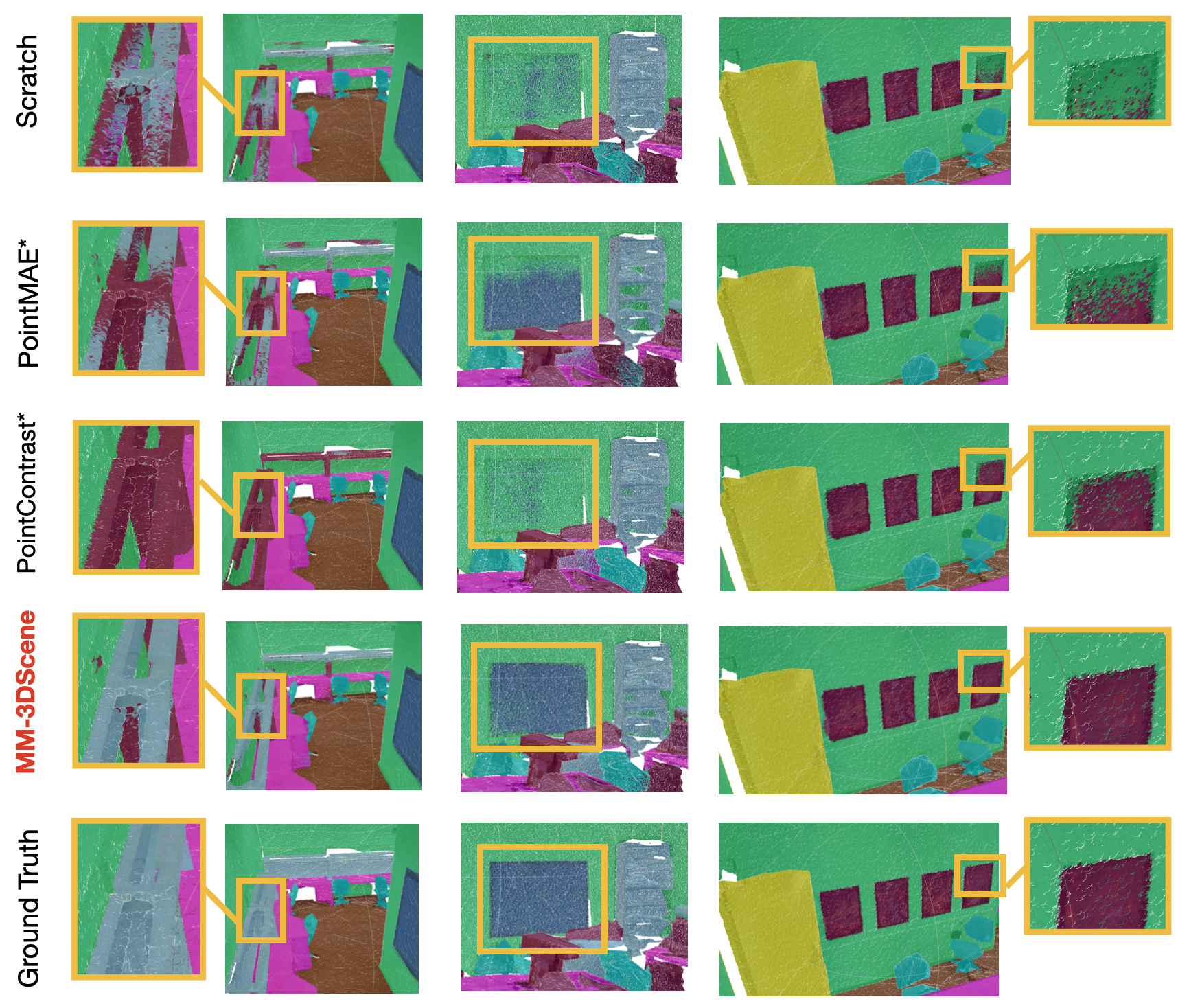}     \caption{Qualitative results on S3DIS semantic segmentation. } 
    \label{fig:vis_semanticseg}
\end{figure}

\begin{figure*}[tb]
    \centering
    \includegraphics[width=1\textwidth]{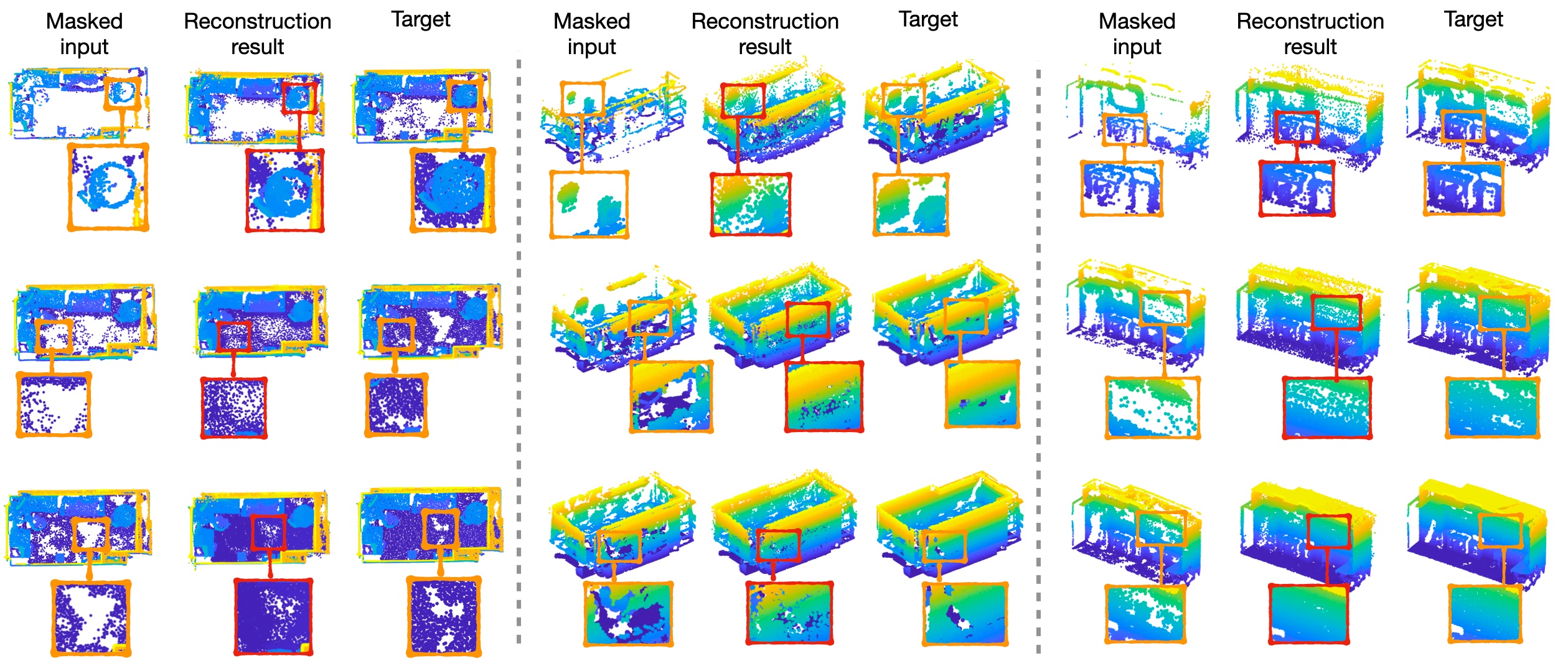}
    \caption{Qualitative results of MM-3DScene masked reconstruction pretext task.
    Our method can effectively reconstruct the masked areas, suggesting it learned rich visual representations for understanding 3D scenes.
    } \label{fig:supple_recon_vis}
\end{figure*}

% Our method explicitly reduces the ambiguity during the masked reconstruction of 3D scenes.
Fig.~\ref{fig:supple_recon_vis} visualizes the masked reconstruction results of MM-3DScene. 
It can be observed that:
\textbf{i)} For masked \textbf{input}, our mask strategy \textit{preserves} Informative Points to provide basic geometric information, which explicitly reduces the ambiguity during masked reconstruction.
\textbf{ii)} For \textbf{target}, instead of being the original intact scene, it is a relatively more complete one with a smaller masking ratio. 
This prompts models to \textit{concentrate} on reconstructing the local regional 3D structures where models focus on recovering \textit{regional} geometric patterns.
% In doing so, it enjoys less ambiguity and is able to restore accurate geometric information.
\textbf{iii)} For reconstruction \textbf{result}, our model is able to recover the masked areas, suggesting it successfully learned numerous visual representations.
For example, our method works well to recover details of the masked foreground objects (\eg table and chair).
For the background surfaces (\eg floor, wall), our method can also achieve a smooth and complete recovery.
% ?than  target with the smaller masking ratio,
% c?onsidering
% \todo{not only the holistic structure but also accurate fine-grained details}
In addition to the visualization of the  pre-training reconstruction results, as shown in the Fig.~\ref{fig:vis_semanticseg}, we present the visualization results of the downstream semantic segmentation task. It can be seen that compared with other methods, our method can correct the results in some areas where the prediction is inaccurate.

\begin{figure}[!htbp] 
\centering
    \includegraphics[width=0.47\textwidth]{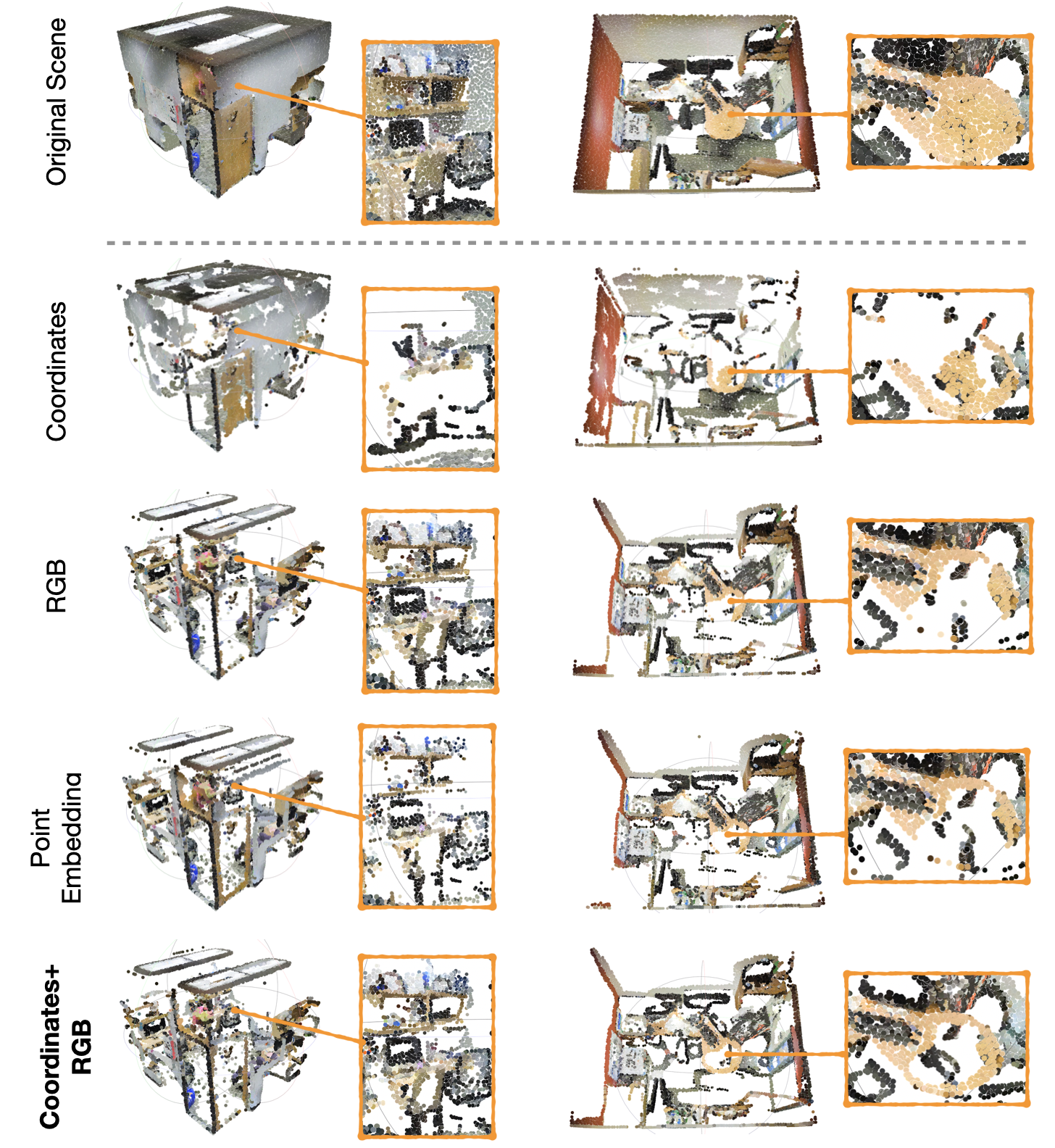}     \caption{Visualization of masked scene guided by different formats of local statistics. 
    In this figure, the \textbf{contour of a table} (\ie, the representative geometric structures) is accurately found and preserved, when calculating the local difference of coordinates+RGB.}
    \label{fig:vis_statistics_mask}
\end{figure}

% \section*{Appendix B: Masked Reconstruction Results}

\section*{Appendix B: More Ablations of Masked Reconstruction}

\paragraph{More formats of local statistics.}
% \paragraph{}
In our Local-Statistics-Guided Masking, we exploit local statistics to discover informative points, aiming to \textit{accurately} retain the representative geometric structures.
The local statistics are denoted by the local difference between each point and its neighboring points in terms of coordinates and colors.
Here we investigate other possible formats of the local difference, including point embedding difference (gained by applying MLPs \cite{mlp} on each point), only coordinates difference, and only RGB difference.
Fig.~\ref{fig:vis_statistics_mask} shows that considering the local difference of both coordinates and RGB is the most applicable way to find informative points. 
For example in this figure, the \textit{structural contour} of a table is \textit{accurately} found and preserved, when guided by the local differences of coordinates+RGB, which provides useful information hints for recovering the masked interior area of this table.
As a result, our method performs best when the masking is guided by the local statistic of coordinates+RGB, as listed in Table~\ref{tab:result_statistics_mask}.

\begin{figure}[tb] 
\centering
    \includegraphics[width=0.35\textwidth]{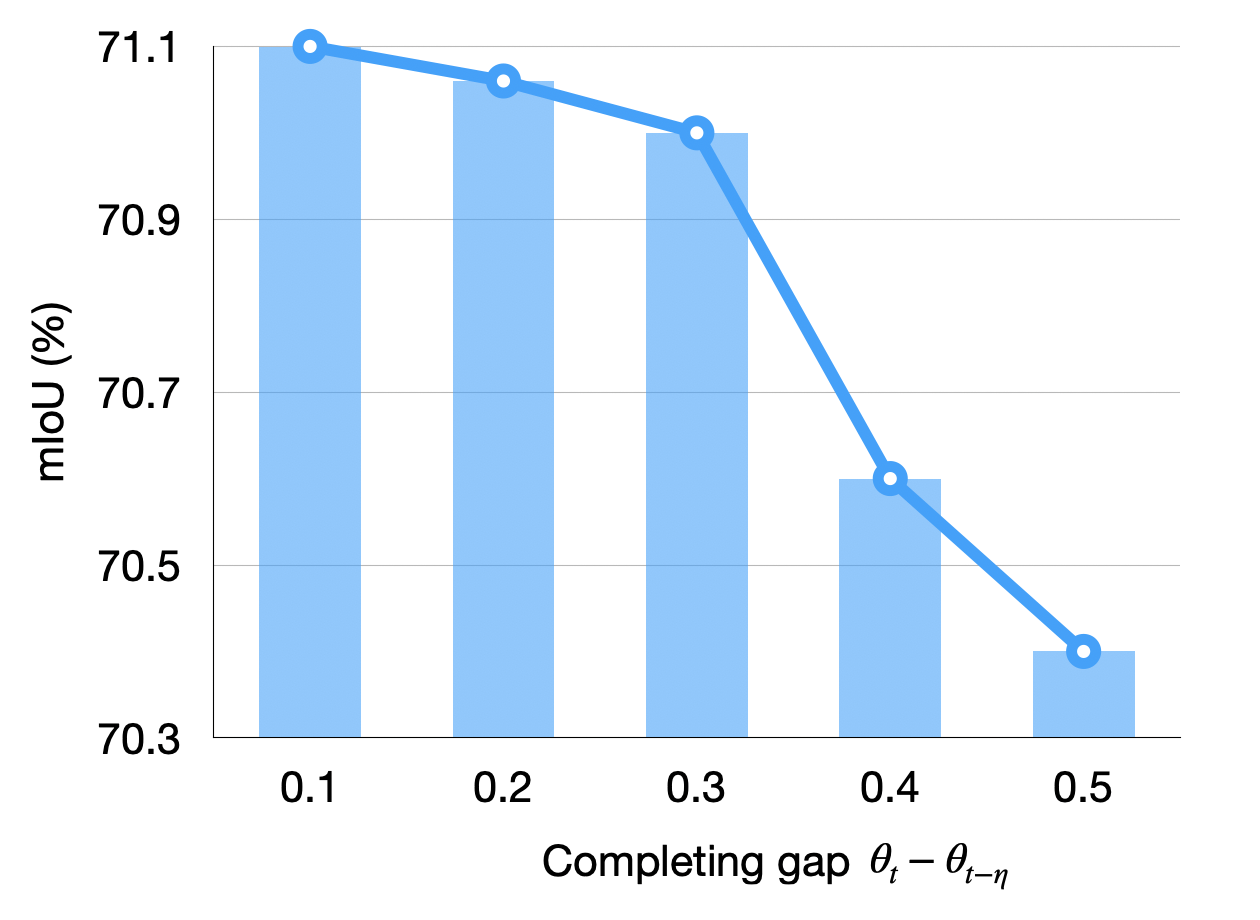}     \caption{Ablation study of \textbf{masked reconstruction gap} on S3DIS semantic segmentation (based on MM-3DScene w/o $L_{CSD}$).} 
    \label{fig:vis_completinggap}
\end{figure}

\paragraph{Ablation studies of reconstruction gap.}
Our method uses the incremental masking ratio $\mathcal{\theta} = \{ \theta_1, ..., \theta_t,...,\theta_T \}$ to progressively mask the scene.
During the masked reconstruction, the masking ratio is $\theta_t$ for the input scene, and $\theta_{t-\eta}$ for the target scene, where $\eta$ indicates the masked gap to be recovered and latently influences the \textit{difficulty} of the pretext task.
Fig.~\ref{fig:vis_completinggap} provides the ablation study of such reconstruction gap, where our model enjoys the least difficulty and performs best under $\theta_t - \theta_{t-\eta}=0.1$, and degrades when the gap becomes larger.
Additionally, we also implement the random reconstruction gap, which probably causes more ambiguity, yielding 70.36\% mIoU.

% \paragraph{Ablation studies of KNN and T}
% \my{
% Since the Local-Statistics-Guided Masking is important for the proposed pretraining.
% Here we conduct the ablation studies on how to select hyperparameters in masking, such as KNN neighboring used, T in scene sequence.
% % The results are summarized in Table \ref{tab:ablation_knn} and \ref{tab:ablation_T}.
% %
% % }

% \begin{table}[t]\centering
%     \resizebox{0.46\textwidth}{!}{
%     \scriptsize
%         \begin{tabular}{l | *{5}{c}}
%                 \toprule
%                 K & 20 & 40 & \textbf{60} & 80 & 100\\
%                 \midrule
%                 mIoU & 70.5 & 70.7 & \textbf{71.1} & 70.9 & 70.9\\
%                 \bottomrule
%             \end{tabular}
%     }
%     \caption{Ablations on KNN of  local statistics guided masking.}
%     \label{tab:ablation_knn}
% \end{table}

% \begin{table}[t]\centering
%     \resizebox{0.46\textwidth}{!}{
%     \scriptsize
% \begin{tabular}{l | *{4}{c}}
%         \toprule
%         T & 5 & 10 & 50 & \textbf{100}\\
%         \midrule
%         mIoU & 70.48 & 70.79 & 71.07 & \textbf{71.10}\\
%         \bottomrule
%     \end{tabular}
%     }
%     \caption{Ablations on T of local statistics guided masking.}
%     \label{tab:ablation_T}
% \end{table}

\begin{table}[t]\centering
    \resizebox{0.47\textwidth}{!}{
    \large
    \begin{tabular}{{c}|{l} {l} *{5}{c}}
        \toprule
      & Pre-training method & Local Statistic  & mIoU & mAcc  \\
        \midrule
      $\romannumeral 1$ & \textcolor{gray}{Scratch} & -  & \textcolor{gray}{70.4} & \textcolor{gray}{76.5} \\
       % $\romannumeral 2$ & Random Mask as \cite{pointmae}  & $\mathcal{L}_{PC}$ 
       % & 70.4 \small\textcolor{gray}{(+0.0)}  & 76.4 \small\textcolor{gray}{(-0.1)}  \\
      $\romannumeral 2$ &  MM-3DScene (w/o $L_{CSD}$) &  Coordinates 
       & 70.6  \small\textcolor{gray}{(+0.2} & 76.5  \small\textcolor{gray}{(+0.0)}\\
      $\romannumeral 3$ &  MM-3DScene (w/o $L_{CSD}$) &  RGB
       & 70.9  \small\textcolor{gray}{(+0.5)} & 77.0  \small\textcolor{gray}{(+0.5)}\\
      $\romannumeral 4$ &  MM-3DScene (w/o $L_{CSD}$) &  point embedding
       & 70.9  \small\textcolor{gray}{(+0.5)} & 76.9  \small\textcolor{gray}{(+0.4)}\\
      $\romannumeral 5$ &  MM-3DScene (w/o $L_{CSD}$) &  \textbf{Coordinates + RGB}
       & \textbf{71.1}  \small\textcolor{gray}{(+0.7)} & \textbf{77.2}  \small\textcolor{gray}{(+0.7)}\\       
        \bottomrule
    \end{tabular}
    }
    \caption{MM-3DScene (w/o $L_{CSD}$) guided by \textbf{different formats of local statistics} for S3DIS semantic segmentation.}
    \label{tab:result_statistics_mask}
\end{table}

\begin{table}[t]\centering
    \resizebox{0.46\textwidth}{!}{
    \scriptsize
    \begin{tabular}{{l}*{10}{c}}
        \toprule
       Method &  \text{Model Size} & Train  Time  & Infer Time  & mAP@0.25 & mAP@0.5\\
        \midrule
       VoteNet \cite{votenet} (scratch) & \text{0.95M} & 5.9h & 0.2s & 58.7 & 35.4\\
       MM3DScene + VoteNet   & \text{1.48M } & 15.6h  & - &  63.1 \scriptsize\textcolor{gray}{(+4.4)} & 41.5 \scriptsize\textcolor{gray}{(+6.1)}\\
        \midrule
        H3DNet \cite{h3dnet} (scratch) & \text{4.74M} & 12.3h & 7.3s & 64.8 & 47.4\\
        MM3DScene + H3DNet  & \text{6.87M} & 36.1h & - & 66.8 \scriptsize\textcolor{gray}{(+2.0)} & 48.9 \scriptsize\textcolor{gray}{(+1.5)}\\
        \bottomrule
    \end{tabular}
    }
    \caption{3D object detection results on ScanNetv2. The baseline results come from official code implementations. The training  and inference times   are evaluated with the same training settings.}
    \label{tab:supple_h3dnet}
\end{table}

\begin{table}[t]\centering
    \resizebox{0.46\textwidth}{!}{
    \scriptsize
    \begin{tabular}{{l}*{10}{c}}
        \toprule
       Method & \text{Model Size}  & Train  Time & Infer Time & mIoU\\
        \midrule
       PointTrans\cite{pointtrans} (scratch) & \text{7.76M} & \text{17.3h} 
 & 4.36s &  70.4\\
       MM3DScene + PointTrans & \text{8.63M} & \text{29.1h} & - & 71.9 \scriptsize\textcolor{gray}{(+1.5)}\\
        \midrule
        Stratified Trans* \cite{lai2022stratified} (scratch) & \text{8.02M} & \text{45.7h} & 11.77s  &  70.3\\
        MM-3DScene + Stratified Trans* &\text{8.89M} & \text{73.2h} &  
- & 71.6\scriptsize\textcolor{gray}{(+1.3)}\\
        \bottomrule
    \end{tabular}
    }
    \caption{3D semantic segmentation results on S3DIS. The baseline results come from official code implementations. The training  and inference times  are evaluated with the same training settings.}
    \label{tab:supple_stformer}
\end{table}

\begin{table*}[t]
    \centering
    \footnotesize% fontsize
    \setlength{\tabcolsep}{6pt}% column separation
    \renewcommand{\arraystretch}{1.2}%row space 
    \begin{tabular}{l|c|ccccccccccccc}%ccccc
        \hline \text {Method} &   \text {mIoU } & \text {ceil.} & \text {floor} & \text {wall} & \text {beam} & \text {col.} & \text {win.} & \text {door} & \text {table} & \text {chair} & \text {sofa} & \text {bookc.} & \text {board} & \text {clu.} \\
        % \text {RSNet} & - & 66.5 & 56.5 & 92.5 & 92.8 & 78.6 & 32.6 & 34.4 & 51.6 & 68.1 & 60.1 & 59.7 & 50.2 & 16.4 & 44.9 & 52.0 \\
        \hline
        \text {PointNet \cite{pointnet}}   & 41.1 & 88.8 & 97.3 & 69.8 & 0.1 & 3.9 & 46.3 & 10.8 & 58.9 & 52.6 & 5.9 & 40.3 & 26.4 & 33.2\\         
        \text{SegCloud \cite{tchapmi2017segcloud} } & 48.9 & 90.1 & 96.1 & 69.9 & 0.0 & 18.4 & 38.4 & 23.1 & 70.4 & 75.9 & 40.9 & 58.4 & 13.0 & 41.6\\
        \text{TangentConv \cite{tatarchenko2018tangent}}  & 52.6 & 90.5 & 97.7 & 74.0 & 0.0 & 20.7 & 39.0 & 31.3 & 77.5 & 69.4 & 57.3 & 38.5 & 48.8 & 39.8\\
        \text {SPGraph \cite{landrieu2018large}}   & 58.0 & 89.4 & 96.9 & 78.1 & 0.0 &  {42.8} & 48.9 & 61.6 &  {84.7} & 75.4 & 69.8 & 52.6 & 2.1 & 52.2 \\
        \text{PCNN \cite{pcnn}}   & 58.3 & 92.3 & 96.2 & 75.9 & {0.3} & 6.0 &  {{69.5}} &  {63.5} & 65.6 & 66.9 & 68.9 & 47.3 & 59.1 & 46.2\\        
        \text{RNNFusion \cite{ye20183d}}  & 57.3 & 92.3 & 98.2 & 79.4 & 0.0 & 17.6 & 22.8 & 62.1 & 80.6 & 74.4 & 66.7  & 31.7& 62.1 &  {56.7}\\
        \text{Eff 3D Conv \cite{zhang2018efficient}}  & 51.8 & 79.8 & 93.9 & 69.0 & 0.2 & 28.3 & 38.5 & 48.3 & 73.6 & 71.1 & 59.2 & 48.7 & 29.3 & 33.1 \\
        \text {PointCNN \cite{pointcnn}}   & 57.3 & 92.3 & 98.2 & 79.4 & 0.0 & 17.6 & 22.8 & 62.1 & 74.4 & 80.6 & 31.7 & 66.7 & 62.1 &  {56.7} \\
        \text {PointWeb \cite{pointweb}}   & 60.3 & 92.0 &  {98.5} & 79.4 & 0.0 & 21.1 & 59.7 & 34.8 & 76.3 &  {88.3} & 46.9 & 69.3 & 64.9 & 52.5 \\
         {IAF-Net\cite{xu2021investigate}} &  {64.6}& 91.4 &  {98.6} & 81.8 & 0.0 & 34.9 &  {62.0} & 54.7 & 79.7 & 86.9 & 49.9 &  {72.4} &  {74.8} & 52.1 \\
        \text{KPConv \cite{kpconv}}  &  {67.1} & {92.8} & 97.3 &  {82.4} & 0.0 & 23.9 & 58.0 & { 69.0} & 81.5 &  {91.0} &  {75.4} &  {75.3} & 66.7 &  {58.9} \\       
        PointTransformer \cite{pointtrans} & 70.4 & 94.0 & 98.5 & 86.3 & 0.0 & 38.0 & 63.4 & 74.3 & 89.1 & 82.4 & 74.3 & 80.2 & 76.0 & 59.3\\
        \midrule
        MM-3DScene(Ours) & 71.9 & 94.6 & 98.6 & 87.1 & 0.0 & 44.2 & 62.9 &
        79.2 & 90.7 & 81.7 & 74.3 & 81.4 & 79.3 & 60.3\\
        \hline
    \end{tabular}
        \caption{ Semantic segmentation results on S3DIS dataset evaluated on Area 5.}
    \label{tab:s3dis_area5_detail}
\end{table*}

% \section{MM-3DScene without RGB Input}
% The local-statistic-guided masking in our method exploits the local difference on both coordinates and RGB colors. Although most 3D datasets include RGB colors, RGB information can not be directly gained by LiDAR sensors for automotive driving datasets \cite{kitti,nuscenes,waymo}.
% To this end, here we investigate our method without RGB input.

% \section{MM-3DScene for Shape-level Task}

\section*{Appendix C: Other Backbones with MM-3DScene}
In the main paper, we adopt VoteNet \cite{votenet} as the backbone for object detection, and Point Transformer \cite{pointtrans} for semantic segmentation.
In this section, we utilize other backbone networks for verifying the generalization ability of our MM-3DScene.

% \subsection{3D Object Detection}
\paragraph{H3DNet object detection.}
We apply our MM-3DScene pretrained framework on H3DNet \cite{h3dnet} which is a more powerful network using hybrid geometric primitives based on VoteNet \cite{votenet}. 
As shown in Table.~\ref{tab:supple_h3dnet}, MM-3DScene improves the H3DNet with the mAP@0.25 by 2.0 and mAP@0.5 by 1.5, exceeding the performance with VoteNet as the backbone.

% \subsection{3D Semantic Segmentation}

\paragraph{Stratified Transformer semantic segmentation.}
We also evaluate the performance of Stratified Transformer \cite{lai2022stratified} as the backbone on S3DIS semantic segmentation. We reproduce the backbone performance using its official code and report the results in Table.~\ref{tab:supple_stformer}. Our MM-3DScene surpasses Stratified Transformer by 1.3\% mIoU. However, it comes with a high computational cost (2.5 times of MM3D-Scene + PT) and a long training time.

% Meanwhile, we also verify the performances of using Stratified Transformer \cite{lai2022stratified} as backbone on S3DIS semantic segmentation,
% We reproduce the  performance of backbone using its official code,
% the results are shown in Table.~\ref{tab:supple_stformer}.
% Our MM-3DScene achieves a performance gain of 1.3\% mIoU over Stratified Transformer. 
% however, it requires a long training time at the cost (2.5 times of MM3D-Scene + PT).

% Very recently, Stratified Transformer \cite{lai2022stratified} optimizes Point Transformer \cite{pointtrans} via enlarging the effective receptive field, and gets superior performance.
% % but its relatively long training time is unacceptable.
% \todo{As shown in Table.~\ref{tab:supple_stformer}, MM-3DScene can further enhance Stratified Transformer for semantic segmentation.}
% However, 

\paragraph{Discussions.}
Although both H3DNet \cite{h3dnet} and Stratified Transformer \cite{lai2022stratified} inherit VoteNet \cite{votenet} and Point Transformer \cite{pointtrans}, and achieve decent performance, they introduce highly-engineered architectures tailored to their network-specific operations, making it difficult to evaluate the improvement made by the self-supervised frameworks.
Thus, we advocate simple and classical baselines, with the goal of minimizing the influence of network architectures to better measure the performance gain \textit{purely} from the self-supervised pretraining framework -- MM-3DScene.

Moreover, both Point Transformer \cite{pointtrans} and VoteNet \cite{votenet} stand out with conspicuously excellent \textbf{efficiency}, as reflected in \textit{model size}, \textit{training time}, and \textit{inference time} of Table~\ref{tab:supple_h3dnet} and Table~\ref{tab:supple_stformer}, which is highly important for the deployment on real applications.

\section*{Appendix D: More fine-grained quantitative results}
To provide a more comprehensive analysis, we present the segmentation results of each category in Table \ref{tab:s3dis_area5_detail}. We observe that most categories have different degrees of improvement over the Point Transformer\cite{pointtrans} backbone that we use. For instance, we achieve 6.2\% gain on column, 4.9\% on door, 3.3\% on board, and slight decrease on window and chair.

\end{document}